\documentclass[10pt,twocolumn,letterpaper]{article}

\usepackage{cvpr}              

\usepackage{graphicx}
\usepackage{amsmath}
\usepackage{amssymb}
\usepackage{booktabs}

\usepackage{mathtools}
\usepackage{xcolor}
\usepackage{enumitem}
\usepackage{algorithm,algpseudocode}
\usepackage{array}
\usepackage{diagbox}
\usepackage{amsfonts}
\usepackage{multirow}
\usepackage{bm}

\usepackage{color, colortbl}

\newtheorem{definition}{Definition}[section]

\usepackage[pagebackref,breaklinks,colorlinks]{hyperref}

\usepackage[capitalize]{cleveref}
\crefname{section}{Sec.}{Secs.}
\Crefname{section}{Section}{Sections}
\Crefname{table}{Table}{Tables}
\crefname{table}{Tab.}{Tabs.}

\usepackage[accsupp]{axessibility}  

\def\cvprPaperID{11402}
\def\confName{CVPR}
\def\confYear{2022}

\DeclareMathOperator{\arccosh}{arccosh}

\newcommand{\tb}[3]{\setlength{\tabcolsep}{#2mm}\begin{tabular}{#1}#3\end{tabular}}

\begin{document}

\title{Clipped Hyperbolic Classifiers Are Super-Hyperbolic Classifiers}

\author{
\tb{@{}cccc@{}}{5}{
Yunhui Guo$^{1}$ & 
Xudong Wang$^{1}$ & 
Yubei Chen$^{2}$ &
Stella X. Yu$^{1}$\\
}\\
\tb{cc}{5}{
$^{1}$UC Berkeley / ICSI&  
$^{2}$Facebook AI Research
}}

\maketitle

\begin{abstract}
Hyperbolic space can naturally embed hierarchies, unlike Euclidean space. Hyperbolic Neural Networks (HNNs) exploit such representational power by lifting Euclidean features into hyperbolic space for classification, outperforming Euclidean neural networks (ENNs) on datasets with known semantic hierarchies. However, HNNs underperform ENNs on standard benchmarks without clear hierarchies, greatly restricting HNNs' applicability in practice.

Our key insight is that HNNs' poorer general classification performance results from vanishing gradients during backpropagation, caused by their hybrid architecture connecting Euclidean features to a hyperbolic classifier. We propose an effective solution by simply clipping the Euclidean feature magnitude while training HNNs.  

Our experiments demonstrate that clipped HNNs become super-hyperbolic classifiers: They are not only consistently better than HNNs which already outperform ENNs on hierarchical data, but also on-par with ENNs on MNIST, CIFAR10, CIFAR100 and ImageNet benchmarks,  with better adversarial robustness and out-of-distribution detection.

\end{abstract}

\section{Introduction}
Many datasets are inherently hierarchical. WordNet \cite{miller1995wordnet} has a hierarchical conceptual structure, users in social networks such as Facebook or twitter form hierarchies based on different occupations and organizations \cite{gupte2011finding}. 

Representing such hierarchical data in Euclidean space cannot capture and reflect their semantic or functional resemblance \cite{alanis2016efficient,nickel2017poincar}. Hyperbolic space, i.e., non-Euclidean space with constant negative curvature, has been leveraged to embed data with hierarchical structures with low distortion owing to the nature of exponential growth in volume with respect to its radius \cite{nickel2017poincar,sarkar2011low,sala2018representation}. For instance, hyperbolic space has been used for analyzing the hierarchical structure in single cell data \cite{klimovskaia2020poincare}, learning hierarchical word embedding \cite{nickel2017poincar}, embedding complex networks \cite{alanis2016efficient}, etc.

Recent algorithms operate directly in hyperbolic space   to exploit more representational power.   Examples are Hyperbolic Perceptron \cite{weber2020robust}, Hyperbolic Support Vector Machine \cite{cho2019large}, and
Hyperbolic Neural Networks (HNNs) \cite{ganea2018hyperbolic}, an alternative to standard Euclidean neural networks (ENNs).  

\begin{figure}[!t]
   \centering
\begin{tabular}{cc}
\multicolumn{2}{c}{ \includegraphics[width=0.45\textwidth]{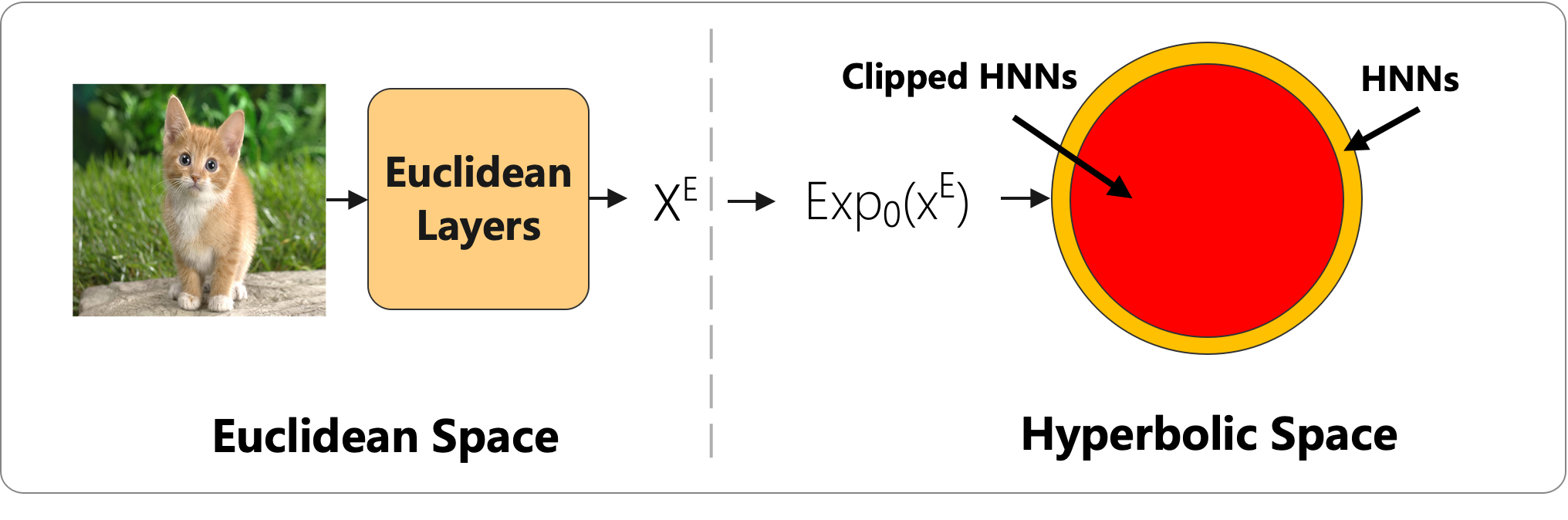}}\\
\multicolumn{2}{c}{a)  HNNs employ a hybrid architecture.}
\\
\includegraphics[width=0.22\textwidth]{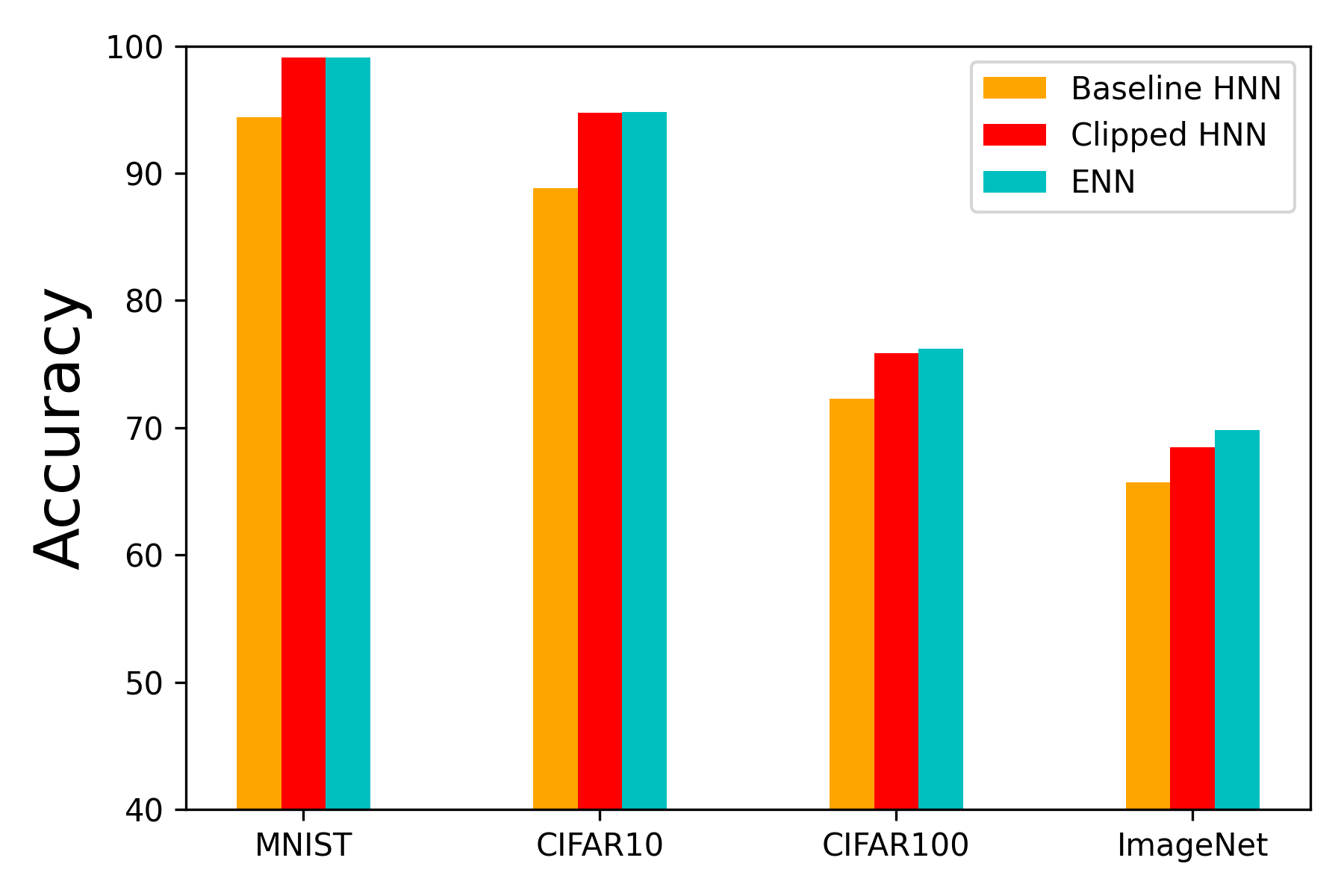}&
\includegraphics[width=0.22\textwidth]{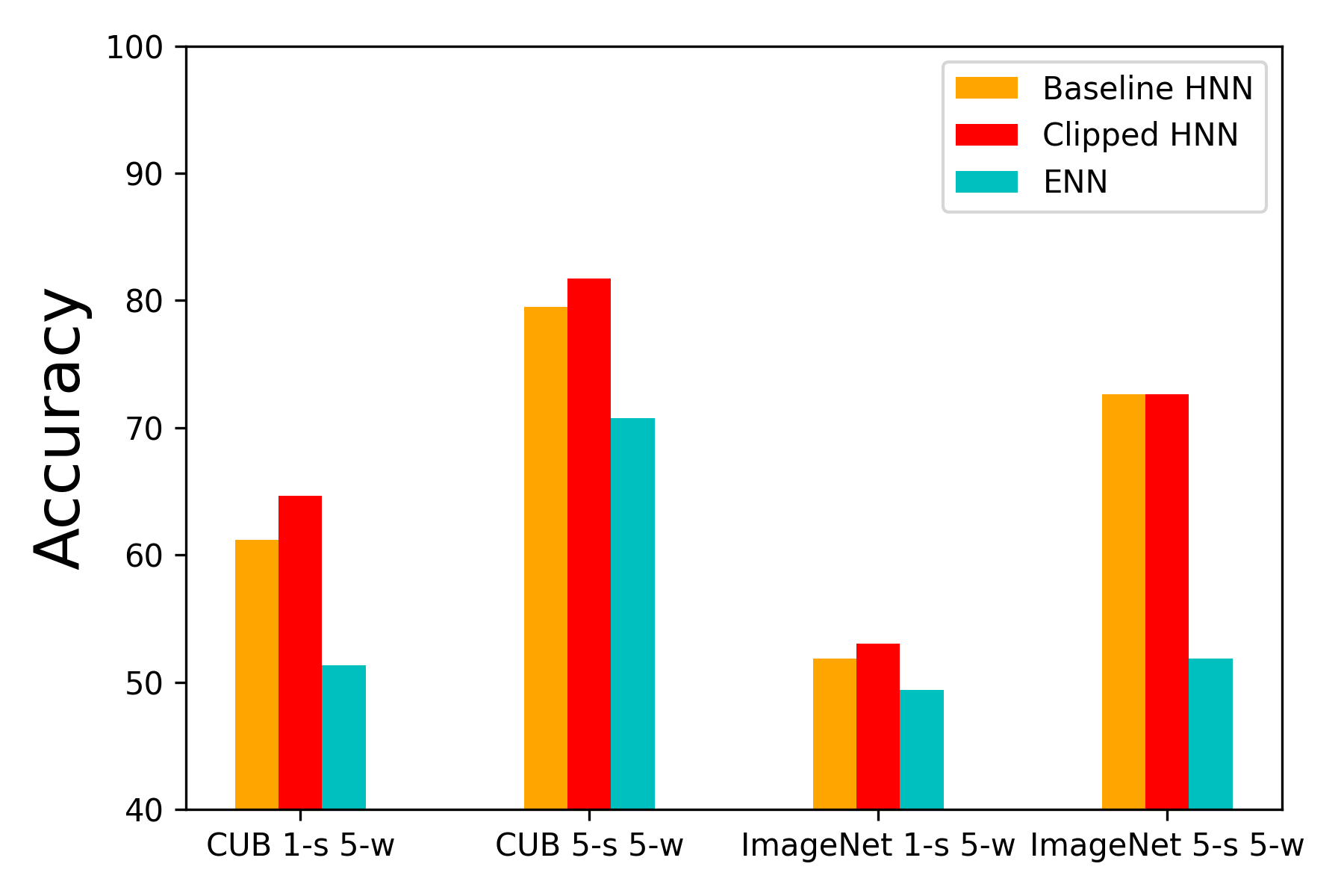}  \\
b) Standard benchmarks.   & c) Few-shot learning tasks.
\end{tabular}
\caption{ We propose an effective solution for training HNNs by clipping the Euclidean features. Clipped HNNs become super-hyperbolic classifiers: They are not only consistently better than HNNs which already outperform ENNs on hierarchical data, but also on-par with ENNs on standard benchmarks. a) HNNs employ a hybrid architecture. The Euclidean part converts an input into Euclidean embedding. Then the Euclidean embedding is projected onto the Poincar\'e model of hyperbolic space via exponential map Exp$_{0}(\cdot)$. Finally, the hyperbolic embeddings are classified with Poincar\'e hyperplanes. Clipped HNNs utilize a reduced region of hyperbolic space. b) Clipped HNNs outperform baseline HNNs on standard benchmarks. c) Clipped HNNs outperform both HNNs and ENNs on 1-s (shot) 1-w (way) and 5-s (shot) 5-w (way) few-shot learning tasks.  }
    \label{fig:HNNs}
\end{figure}

HNNs adopt a hybrid architecture \cite{khrulkov2020hyperbolic} (Figure \ref{fig:HNNs}): An ENN is first used for extracting image features in Euclidean space; they are then projected onto hyperbolic space to be classified by a hyperbolic multiclass logistic regression \cite{ganea2018hyperbolic}.

While HNNs outperform ENNs on several datasets with explicit hierarchies \cite{ganea2018hyperbolic}, there are several serious limitations.  {\bf 1)}
HNNs underperform ENNs on standard classification benchmarks with flat or non-hierarchical semantic structures.
{\bf 2)} Even for image datasets that possess latent hierarchical structures 
there are no experimental evidence  that HNNs can capture such structures or provide on-par performance with ENNs \cite{khrulkov2020hyperbolic}.
{\bf 3)} Existing improvements on HNNs mainly focus on reducing the number of parameters \cite{shimizu2020hyperbolic} or incorporating different types of neural network layers such as attention \cite{gulcehre2018hyperbolic} or convolution \cite{shimizu2020hyperbolic}.  Unfortunately, why HNNs are worse than ENNs on standard benchmarks has not been investigated or understood.

Our key insight is that HNNs' poorer general classification performance is caused by their hybrid architecture connecting Euclidean features to a hyperbolic classifier.  It leads to \emph{vanishing gradients} during training. In particular, the training dynamics of HNNs push the hyperbolic embeddings to the boundary of the Poincar\'e ball \cite{anderson2006hyperbolic} which causes the gradients of Euclidean parameters to vanish.  

We propose a simple yet effective solution to this problem by simply clipping the Euclidean feature magnitude during training, thereby preventing the hyperbolic embedding from approaching the boundary during training. Our experiments demonstrate that clipped HNNs become super-hyperbolic classifiers: They are not only consistently better than HNNs which already outperform ENNs on hierarchical data, but also on-par with ENNs on MNIST, CIFAR10, CIFAR100 and ImageNet benchmarks, with better adversarial robustness and out-of-distribution detection.

Our paper makes the following contributions. {\bf 1)} Our detailed analysis reveals the underlying issue of vanishing gradients that makes HNNs worse than ENNs on standard classification benchmarks. {\bf 2)} We propose a simple yet effective feature clipping solution. {\bf 3)} Our extensive experimentation demonstrates that clipped HNNs outperform standard HNNs and become on-par with ENNs on standard benchmarks. They are also more robust to adversarial attacks and exhibit stronger out-of-distribution detection capability than their Euclidean counterparts.

\section{Related Work}

\noindent \textbf{Supervised Learning in Hyperbolic Space.}  Several hyperbolic neural networks were proposed in the seminal work of HNNs \cite{ganea2018hyperbolic},
including multinomial logitstic regression (MLR), fully connected and recurrent neural networks which can operate directly on hyperbolic embeddings, outperforming Euclidean variants on text entailment and noisy-prefix prediction tasks. Hyperbolic Neural Networks++ \cite{shimizu2020hyperbolic} reduces the number of parameters of HNNs and also introduces hyperbolic convolutional layers. Hyperbolic attention networks \cite{gulcehre2018hyperbolic} rewrite the operations in the attention layers using gyrovector operations \cite{ungar2005analytic}, delivering gains on neural machine translation, learning on graphs and visual question answering.  Hyperbolic graph neural network \cite{liu2019hyperbolic} extends the representational geometry of Graph Neural Networks (GNNs) \cite{zhou2020graph} to hyperbolic space. Hyperbolic graph attention networks \cite{zhang2019hyperbolic} further studies GNNs with attention mechanisms in hyperbolic space.  HNNs have been used for few-shot classification and person re-identification \cite{khrulkov2020hyperbolic}.

\noindent \textbf{Unsupervised Learning in Hyperbolic Space.}   \cite{nagano2019wrapped} uses a wrapped normal distribution in hyperbolic space to construct hyperbolic variational autoencoders (VAEs) \cite{kingma2013auto}, whereas \cite{mathieu2019continuous} uses Gaussian generalizations in hyperbolic space to construct Poincar\'e VAEs. \cite{hsu2020learning} applies hyperbolic neural networks to unsupervised 3D segmentation of complex volumetric data.

Our work differs from all the above-mentioned methods which focus on the application of HNNs to data with natural tree structures. We extend HNNs to standard classification benchmarks which may not have hierarchies. By improving HNNs to the level of ENNs in these scenarios, we greatly enhance the general applicability of HNNs.  

\section{Super-Hyperbolic Classifiers from Clipping}
Our goal is to understand why HNNs underperform ENNs on standard image classification benchmarks and propose corresponding solutions. 
First, we review the basics of Riemannian geometry and HNNs. Then, we analyze the vanishing gradient problem in training HNNs. Finally, we present the proposed method and discuss its connections to existing methods. 

\subsection{Preliminaries}

\noindent  \textbf{Smooth Manifold.} An $n$-dimensional topological manifold $\mathcal{M}$ is a topological space that is locally Euclidean of dimension $n$: Every point $\mathbf{x} \in \mathcal{M}$ has a neighborhood that is homeomorphic to an open subset of $\mathbb{R}^n$. A smooth manifold is a topological manifold with additional smooth structure which is a maximal smooth atlas.

\noindent  \textbf{Riemannian Manifold.} A Riemannian manifold $(\mathcal{M}, \mathfrak{g})$ is a real smooth manifold with a Riemannian metric $\mathfrak{g}$. The Riemannian metric $\mathfrak{g}$ is defined on the tangent space $T_\mathbf{x}{\mathcal{M}}$ of $\mathcal{M}$ which is a smoothly varying inner product. 

\noindent  \textbf{Inner Product and Norm on Riemannian Manifold.} For $\mathbf{x} \in \mathcal{M}$ and any two vectors $ \mathbf{v}, \mathbf{w} \in T_\mathbf{x}{\mathcal{M}}$, the inner product  $\langle \mathbf{v}, \mathbf{w} \rangle_\mathbf{x}$ is defined as $\mathfrak{g}(\mathbf{v}, \mathbf{w})$. With the definition of inner product, for $ \mathbf{v} \in T_\mathbf{x}{\mathcal{M}}$, the norm is defined as $\lVert \mathbf{v} \rVert_\mathbf{x} = \sqrt{ \langle \mathbf{v}, \mathbf{v} \rangle}_\mathbf{x}$. 

\noindent  \textbf{Geodesics on Riemannian Manifold.}  A geodesic is a curve $\gamma : [0, 1] \rightarrow \mathcal{M}$ of constant speed that is locally minimizing the distance between two points on the manifold.

\noindent  \textbf{Exponential Map on Riemannian Manifold.} Given $\mathbf{x}, \mathbf{y} \in \mathcal{M}, \mathbf{v} \in T_\mathbf{x}{\mathcal{M}}$, and a geodesic $\gamma$ of length $\lVert \mathbf{v} \rVert$ such that $\gamma(0) = \mathbf{x}, \gamma(1) = \mathbf{y}, \gamma'(0) =  \mathbf{v} / \lVert  \mathbf{v} \rVert$, the exponential map $\textnormal{Exp}_\mathbf{x}: T_\mathbf{x}{\mathcal{M}} \rightarrow \mathcal{M}$ satisfies $\textnormal{Exp}_\mathbf{x}(\mathbf{v}) = \mathbf{y}$ and the inverse exponential map $\textnormal{Exp}^{-1}_\mathbf{x}: \mathcal{M} \rightarrow T_\mathbf{x}{\mathcal{M}}$ satisfies $\textnormal{Exp}^{-1}_\mathbf{x}(\mathbf{y}) = \mathbf{v}$.  For more details, please refer to \cite{carmo1992riemannian,lee2018introduction}

\noindent \textbf{Poincar\'e Ball Model for Hyperbolic Space.} A hyperbolic space is a Riemannian manifold with constant negative curvature. There are several isometric models for hyperbolic space, one of the commonly used models is Poincar\'e ball model \cite{nickel2017poincar,ganea2018hyperbolic} which can be derived using stereoscopic projection of the hyperboloid model \cite{anderson2006hyperbolic}. The $n$-dimensional Poincar\'e  ball model of constant negative curvature $-c$ is defined as $(\mathbb{B}^n_c, \mathfrak{g}_\mathbf{x}^c)$, where $\mathbb{B}^n_c$ = $\{\mathbf{x} \in \mathbb{R}^n: c\lVert \mathbf{x} \rVert < 1 \}$ and $\mathfrak{g}_\mathbf{x}^c  = (\gamma_\mathbf{x}^c)^2 I_n $ is the Riemannian metric tensor. $I_n$ is the Euclidean metric tensor. The conformal factor is defined as,

\begin{equation}
    \gamma_\mathbf{x}^c = \frac{2}{1- c \lVert \mathbf{x} \rVert^2}
\end{equation}

\noindent The conformal factor induces the inner product $\langle \mathbf{u}, \mathbf{v} \rangle_\mathbf{x}^c = (\gamma_\mathbf{x}^c)^2 \langle \mathbf{u}, \mathbf{v} \rangle$ and norm $\lVert \mathbf{v} \rVert^c_\mathbf{x} = \gamma_\mathbf{x}^c \lVert \mathbf{v} \rVert$ for all $\mathbf{u}, \mathbf{v} \in T_\mathbf{x} \mathbb{B}^n_c$. The exponential map of Poincar\'e ball model can be written analytically with the operations of gyrovector space.

\begin{figure}[!t]
   \centering
\begin{tabular}{c}
\includegraphics[width=0.4\textwidth]{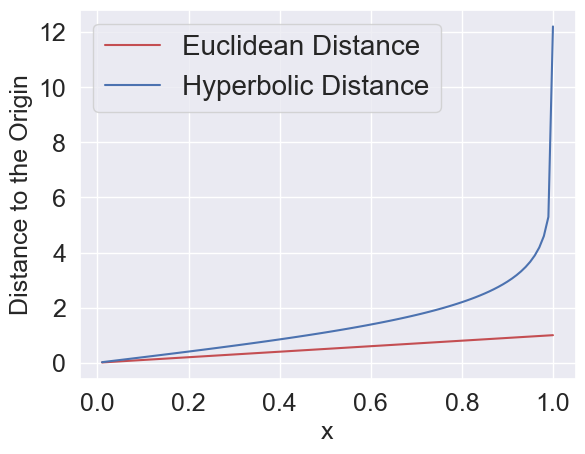}
\end{tabular}
\caption{Hyperbolic distance grows exponentially as we move towards the boundary of the Poincar\'e ball.}
    \label{fig:dist_comparison}
\end{figure}

\begin{figure}[!t]
   \centering
\begin{tabular}{c}
\includegraphics[width=0.25\textwidth]{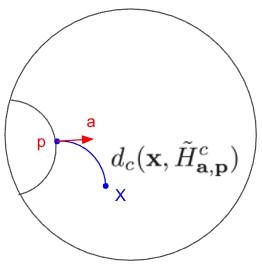}
\end{tabular}
\caption{Poincar\'e hyperplane defined by $\mathbf{a}$ and $\mathbf{p}$. Blue line is the orthogonal projection of $\mathbf{x}$ to the Poincar\'e hyperplane.  }
    \label{fig:p_hyperplane}
\end{figure}

\noindent \textbf{Distance in Poincar\'e Ball Model.} Figure \ref{fig:dist_comparison} shows that hyperbolic distance grows much faster than Euclidean distance as we move towards the boundary of the Poincar\'e ball. As we will show later, this fundamental fact would lead to an optimization issue when we construct a neural network consisting of both Euclidean and hyperbolic layers.

\noindent \textbf{Gyrovector Space.}
\label{sec:gyro} 
A gyrovector space  \cite{ungar2005analytic,ungar2008gyrovector} is an algebraic structure that provides an analytic way to operate in hyperbolic space. Gyrovector space can be used to define various operations such as scalar multiplication, subtraction, exponential map, inverse exponential map in Poincar\'e ball model.

The basic operation in gyrovector space is called M\"{o}bius addition $\oplus_c$. With M\"{o}bius addition $\oplus_c$, we can define vector addition of two points in Poincar\'e ball model as,
\begin{equation}
    \mathbf{u} \oplus_c \mathbf{v} = \frac{ (1+2c \langle \mathbf{u}, \mathbf{v} \rangle + c \lVert \mathbf{v} \rVert^2) \mathbf{u}  + (1 - c \lVert \mathbf{u} \rVert^2)\mathbf{v}   }{  1 + 2c \langle \mathbf{u}, \mathbf{v} \rangle + c^2 \lVert \mathbf{u} \rVert^2 \lVert \mathbf{v} \rVert^2 }
\end{equation}
for all $\mathbf{u}, \mathbf{v} \in \mathbb{B}^n_c$. Particularly, $\lim_{c \rightarrow 0}\oplus_c $ converges to the standard $+$ in the Euclidean space. For more details, please refer to the Supplementary.
 
\noindent \textbf{Hyperbolic Neural Networks.} Hyperbolic neural networks consist of an Euclidean sub-network and a hyperbolic classifier (Figure \ref{fig:HNNs}). The Euclidean sub-network $E(\mathbf{x})$ converts an input $\mathbf{x}$ such as an image into a representation $\mathbf{x}^E$ in Euclidean space. $\mathbf{x}^E$ is then projected onto hyperbolic space $\mathbb{B}^n_c$ via an exponential map $\textnormal{Exp}^c_{\mathbf{0}}(\cdot)$ as $\mathbf{x}^H \in \mathbb{B}^n_c$. The hyperbolic classifier $H(\mathbf{x}^H)$ performs classification based on $\mathbf{x}^H$ with the standard cross-entropy loss $\ell$.

Let the parameters of the Euclidean sub-network be $\mathbf{w}^E$ and the parameters of the hyperbolic classifier be $\mathbf{w}^H$. Given the loss function $\ell$, the optimization problem can be formalized as,
\begin{equation}
    \min_{\mathbf{w}^E, \mathbf{w}^H} \ell(H( \textnormal{Exp}_\mathbf{0}^c(( E(\mathbf{x}; \mathbf{w}^E)); \mathbf{w}^H), y)
    \label{eq: opt_f}
\end{equation}
where the outer and inner functions are $H: \mathbb{B}^n_c \rightarrow \mathbb{R}$ and $E: \mathbb{R}^m \rightarrow \mathbb{R}^n$. As shown in \cite{ganea2018hyperbolic}, the exponential map is defined as,
\begin{equation}
    \textnormal{Exp}_\mathbf{0}^c(\mathbf{v}) = \tanh (\sqrt{c} \lVert \mathbf{v} \rVert) \frac{\mathbf{v} }{ \sqrt{c} \lVert \mathbf{v}  \rVert}
\end{equation}

The construction of hyperbolic classifier relies on the following definition of Poincar\'e hyperplanes,
\begin{definition}[Poincar\'e hyperplanes  \cite{ganea2018hyperbolic}]
For $\mathbf{p} \in \mathbb{B}^n_c$, $\mathbf{a} \in T_\mathbf{p}\mathbb{B}^n_c \setminus \{\mathbf{0}\}$, the Poincar\'e hyperplane is defined as,
\begin{equation}
    \Tilde{H}^c_{\mathbf{a},\mathbf{p}} \coloneqq \{ \mathbf{x} \in \mathbb{B}^n_c: \langle -\mathbf{p}\oplus_c \mathbf{x}, \mathbf{a} \rangle = 0\}
\end{equation}
where $\mathbf{a}$ is the normal vector. Figure \ref{fig:p_hyperplane} shows the Poincar\'e hyperplane defined by $\mathbf{a}$ and $\mathbf{p}$. 
\end{definition}

\cite{ganea2018hyperbolic} shows that in hyperbolic space the probability that a given $\mathbf{x} \in \mathbb{B}^n_c$ is classified as class $k$ is,
\begin{equation}
\small
    p(y=k|\mathbf{x}) \propto \exp  ( \langle  - \mathbf{p}_k \oplus_c \mathbf{x}, \mathbf{a}_k \rangle) \sqrt{ \mathfrak{g}^c_{\mathbf{p_k}} (\mathbf{a}_k, \mathbf{a}_k )} d_c(\mathbf{x}, \Tilde{H}^c_{\mathbf{a}_k, \mathbf{p}} )
    \label{eq: mlr}
\end{equation}
where $d_c(\mathbf{x}, \Tilde{H}^c_{\mathbf{a}_k, \mathbf{p}})$ is the distance of the embedding $\mathbf{x}$ to the Poincar\'e hyperplane of class $k$ as shown in Figure \ref{fig:p_hyperplane}. In hyperbolic classifier, the parameters are the vectors $\{\mathbf{p}_k\}$ for each class $k$.

\subsection{Vanishing Gradient Problem in Training Hyperbolic Neural Networks}

\noindent \textbf{Training Hyperbolic Neural Networks with Backpropagation.} The standard backpropagation algorithm \cite{rumelhart1986learning} is used for training HNNs \cite{ganea2018hyperbolic,khrulkov2020hyperbolic}. During backpropagation, the gradient of the Euclidean parameters $\mathbf{w}^E$ can be computed as,
\begin{equation}
    \frac{\partial \ell}{\partial \mathbf{w}^E} = (\frac{\partial \mathbf{x}^H}{\partial \mathbf{w}^E})^T  
    \frac{\partial \ell}{\partial \mathbf{x}^H}  
    \label{eq:grad_w_E}
\end{equation}
 where $\mathbf{x}^H$ is the hyperbolic embedding of the input $\mathbf{x}$, $\frac{\partial \mathbf{x}^H}{\partial \mathbf{w}^E}$ is the Jacobian matrix and $\frac{\partial \ell}{\partial \mathbf{x}^H}$ is the gradient of the loss function with respect to the hyperbolic embedding $\mathbf{x}^H$. It is noteworthy that since $\mathbf{x}^H$ is an embedding in hyperbolic space, $\frac{\partial \ell}{\partial \mathbf{x}^H} \in T_{\mathbf{x}^H} \mathbb{B}^n_c$ is the Riemannian gradient \cite{bonnabel2013stochastic} and
\begin{equation}
    \frac{\partial \ell}{\partial \mathbf{x}^H} = 
    \frac{(1-\lVert  \mathbf{x}^H\rVert^2)^2   }{4} \nabla \ell(\mathbf{x}^H)
        \label{eq:inverse}
\end{equation}
where $\nabla \ell(\mathbf{x}^H)$ is the Euclidean gradient.

\begin{figure}
   \centering
\begin{tabular}{c}
\includegraphics[width=0.32\textwidth]{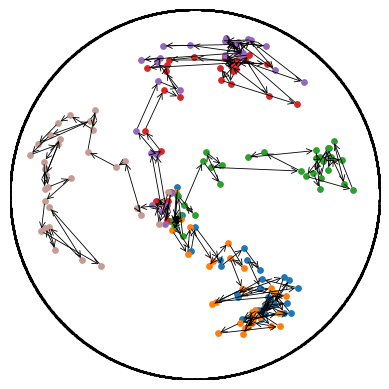} \\
a) \\
\includegraphics[width=0.32\textwidth]{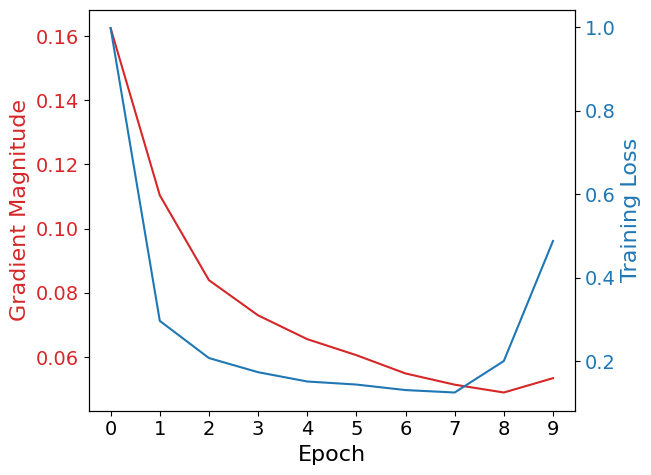} \\
b)
\end{tabular}
    \caption{Hyperbolic neural networks suffer from vanishing gradient problem during training with backpropagation. a) The trajectories of the hyperbolic embeddings of six randomly sampled inputs during training in a 2-dimensional Poincar\'e ball. The arrows indicate the change of location of each embedding with each gradient update. The embeddings move to the boundary of the ball during optimization which causes vanishing gradient problem. b) The gradient vanishes while the training loss goes up at the end of training.}
    \label{fig:vanishing_gradient}
\end{figure}

\noindent \textbf{Vanishing Gradient Problem.} 
We conduct an experiment to show the vanishing gradient problem during training hyperbolic neural networks. We train a LeNet-like convolutional neural network \cite{lecun1998gradient} with hyperbolic classifier on the MNIST data. We use a two-dimensional Poincar\'e ball for visualization. Figure \ref{fig:vanishing_gradient} shows the trajectories of the hyperbolic embeddings of six randomly sampled inputs during training. The arrows indicate the movement of each embedding after one gradient update step. It can be observed that at initialization all the hyperbolic embeddings are close to the center of the Poincar\'e ball. During training, the hyperbolic embeddings gradually move to the boundary of the ball. The magnitude of the gradient diminishes during training as the training loss decays. However, at the end of training, while the training loss slightly increases, the gradient vanishes due to the issue that the hyperbolic embeddings approach the boundary of the ball.

From Equation \ref{eq: mlr}, we can see that in order to maximize the probability of the correct prediction, we need to increase the distance of the hyperbolic embedding to the corresponding Poincar\'e hyperplane, i.e., $ d_c(\mathbf{x}^H, \Tilde{H}^c_{\mathbf{a}_k, \mathbf{p}})$. The training dynamics of HNNs thus push the hyperbolic embeddings to the boundary of the Poincar\'e ball in which case $\lVert \mathbf{x}^H \rVert^2$ approaches one. The inverse of the Riemannian metric tensor becomes zero which causes $\lVert \frac{\partial \ell}{\partial \mathbf{x}^H}\rVert^2$ to be small. From Equation \ref{eq:grad_w_E}, it is easy to see that if $\lVert \frac{\partial \ell}{\partial \mathbf{x}^H}\rVert^2$ is small, then $\lVert \frac{\partial \ell}{\partial \mathbf{w}^E}\rVert^2$ is small and the optimization makes no progress on $\mathbf{w}^E$.

\setlength{\textfloatsep}{1.5pt}

\emph{Vanishing gradient problem} \cite{hochreiter1998vanishing,pennington2017resurrecting,pennington2018emergence,hanin2018neural} is one of the difficulties in training deep neural networks using backpropagation. Vanishing gradient problem occurs when the magnitude of the gradient is too small for the optimization to make progress. For Euclidean neural networks, vanishing gradient problem can be alleviated by architecture designs \cite{hochreiter1997long,he2016deep}, proper weight initialization \cite{mishkin2015all} and carefully chosen activation functions \cite{xu2015empirical}. However, the vanishing gradient problem in training HNNs is not exploited in existing literature.

\noindent \textbf{The Effect of Gradient Update of Euclidean Parameters on the Hyperbolic Embedding.} We derive the effect of a single gradient update of the Euclidean parameters on the hyperbolic embedding, for more details please refer to the Supplementary. For the Euclidean sub-network $E: \mathbb{R}^m \rightarrow \mathbb{R}^n$, consider the first-order Taylor-expansion with a single gradient update, 
\setlength{\belowdisplayskip}{0pt} \setlength{\belowdisplayshortskip}{0pt}
\setlength{\abovedisplayskip}{0pt} \setlength{\abovedisplayshortskip}{0pt}
\begin{equation}
\begin{split}
 E(\mathbf{w}^E_{t+1}) & = E(\mathbf{w}_t^E + \eta \frac{\partial \ell}{\partial \mathbf{w}^E})  \\
 & \approx   E(\mathbf{w}^E_{t}) +   \eta (\frac{\partial E(\mathbf{w}_t^E)}{\partial \mathbf{w}_t^E})^T \frac{\partial \ell}{\partial \mathbf{w}^E} \\
 \end{split}
\end{equation}
where $\eta$ is the learning rate. The gradient of the exponential map can be computed as,
\begin{equation}
\small
\begin{split}
    \nabla \textnormal{Exp}_\mathbf{0}^c(\mathbf{v}) & = \frac{\mathbf{v}}{\sqrt{c} \lVert  \mathbf{v} \rVert} \nabla \tanh({\sqrt{c}\lVert \mathbf{v} \rVert}) + \tanh({\sqrt{c}\lVert \mathbf{v} \rVert}) \nabla \frac{\mathbf{v}}{\sqrt{c} \lVert  \mathbf{v} \rVert} \\
    & = 1-\tanh^2({\sqrt{c}\lVert \mathbf{v} \rVert}) + \tanh({\sqrt{c}\lVert \mathbf{v} \rVert})\frac{1}{\sqrt{c}}  \frac{2}{\lVert \mathbf{v} \rVert}
\end{split}
\end{equation}
Let $\mathbf{x}_{t+1}^H$ be the projected point in hyperbolic space, i.e., 

\begin{equation}
    \mathbf{x}_{t+1}^H = \textnormal{Exp}_\mathbf{0}^c(E(\mathbf{w}^E_{t+1}))
\end{equation}

By applying the first-order Taylor-expansion on the exponential map and following standard derivations, we can find that, 
\begin{equation}
\begin{split}
     \mathbf{x}_{t+1}^H & = \mathbf{x}_{t}^H + C(E(\mathbf{w}^E_{t})^T)\frac{\partial \ell}{\partial \mathbf{w}^E}  \\
\end{split}
\end{equation}

where  $C(E(\mathbf{w}^E_{t})) = \nabla \textnormal{Exp}_\mathbf{0}^c(E(\mathbf{w}^E_{t}))^T\eta (\frac{\partial E(\mathbf{w}_t^E)}{\partial \mathbf{w}_t^E})^T$. Thus once $\lVert  \mathbf{x}_t^H \rVert $ approaches one, from Equation \ref{eq:grad_w_E} and Equation \ref{eq:inverse} we can find that the hyperbolic embedding stagnates no matter how large the training loss is.

\subsection{Clipped Hyperbolic Neural Networks}

\noindent \textbf{Euclidean Feature Clipping.} There are several possible solutions to address the vanishing gradient problem for training HNNs. One tentative solution is to replace all the Euclidean layers with hyperbolic layers, however it is not clear how to directly map the original input images onto hyperbolic space. Another solution is to use normalized gradient descent \cite{hazan2015beyond} for optimizing the Euclidean parameters to reduce the effect of gradient magnitude. However we observed that this introduces instability during training and makes it harder to tune the learning rate for optimizing Euclidean parameters. 

We address the vanishing gradient problem by first reformulating the optimization problem in Equation \ref{eq: opt_f} with a regularization term which controls the magnitude of hyperbolic embeddings,
\begin{equation}
    \min_{\mathbf{w}^E, \mathbf{w}^H} \ell(H( \mathbf{x}^H ; \mathbf{w}^H), y) + \beta \lVert \mathbf{x}^H \rVert^2
    \label{eq: soft}
\end{equation}
where $\mathbf{x}^H = \textnormal{Exp}_\mathbf{0}^c(( E(\mathbf{x}; \mathbf{w}^E))$ and $\beta > 0$ is a hyperparameter. By minimizing the training loss, the hyperbolic embeddings tend to move to the boundary of the Poincar\'e ball which causes the vanishing gradient problem. The additional regularization term is used to prevent the hyperbolic embeddings from approaching the boundary.

\begin{figure}
    \centering
    \includegraphics[width=0.4\textwidth]{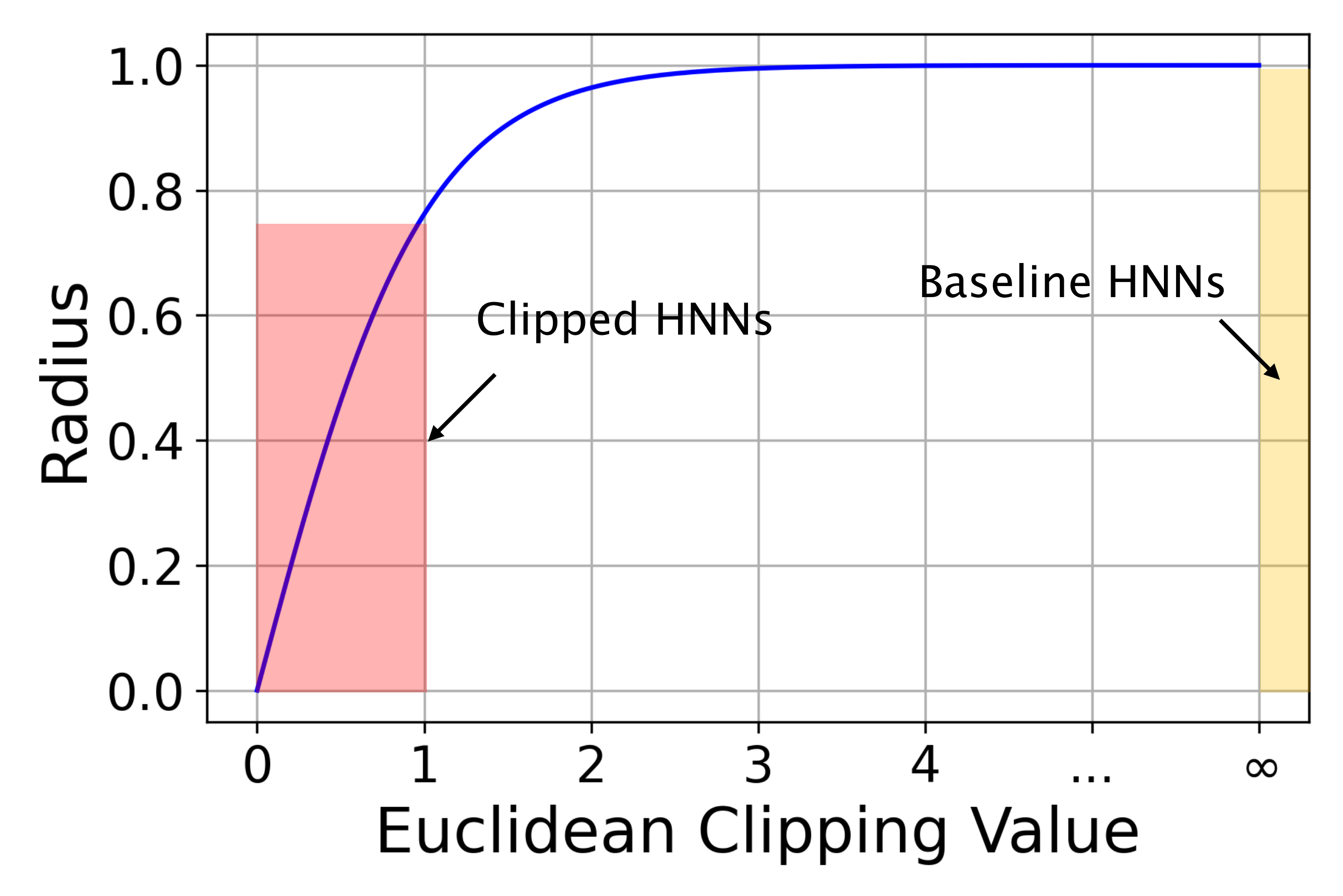}
    \caption{The relation between the clipping value $r$ and the effective radius of the Poincar\'e ball, i.e., $\textnormal{Exp}_\mathbf{0}^c(\textnormal{CLIP}(\mathbf{x}^E; r))$. Clipped HNNs utilize a reduced region of Poincar\'e ball. }
    \label{fig:r_to_h}
\end{figure}

While the soft constraint introduced in Equation \ref{eq: soft} is effective, it introduces additional complexity to the optimization process and has worse performance. We instead employ the following hard constraint which regularizes the Euclidean embedding before the exponential map whenever its norm exceeds a given threshold,
\begin{equation}
    \textnormal{CLIP}(\mathbf{x}^E; r) = \min \{ 1, \frac{r}{\lVert \mathbf{x}^E \rVert} \} \cdot \mathbf{x}^E
\end{equation}
where $\mathbf{x}^E = E(\mathbf{x}; \mathbf{w}^E)$ and  $r > 0$ is a hyperparameter. The clipped Euclidean embedding is projected via exponential map as $\textnormal{CLIP}(\mathbf{x}^H; r) = \textnormal{Exp}_\mathbf{0}^c(\textnormal{CLIP}(\mathbf{x}^E; r))$. The relation between the clipping value and the effective radius of Poincar\'e ball is depicted in Figure \ref{fig:r_to_h}.

Using the relation between the hyperbolic distance and the Euclidean distance to the origin,
\begin{equation}
    d_c(0, x) = s \ln (\frac{s + x}{ s - x})
\end{equation}
where $s = 1 / \sqrt{|c|}$, $c$ is the curvature and $x$ is Euclidean distance to the origin. $c$ is usually set to -1. The clipping value $r$ can be further converted into hyperbolic radius $r_{\mathbb{B}^n_c}$ as below,

\begin{equation}
r_{\mathbb{B}^n_c} = d_c(0, \textnormal{CLIP}(\mathbf{x}^H; r)) = 2r 
\end{equation}

The hyperbolic radius $r_{\mathbb{B}^n_c}$ is measured in hyperbolic distance. The hyperbolic embeddings are within a hyperbolic ball of radius of $r_{\mathbb{B}^n_c}$.

\noindent \textbf{Discussion on Feature Clipping.} The proposed \emph{Feature Clipping} imposes a hard constraint on the maximum norm of the hyperbolic embedding to prevent the inverse of the Riemannian metric tensor from approaching zero. Therefore there is always a gradient signal for optimizing the hyperbolic embedding. Although decreasing the norm of the hyperbolic embedding shrinks the effective radius of the embedding space, we found that it does no harm to accuracy while alleviating the vanishing gradient problem. 

A radius limited hyperbolic classifier is a super-hyperbolic classifier, not a nearly Euclidean classifier. In the Supplementary, we show that clipped hyperbolic space well maintains the hyperbolic property and delivers better results for learning hierarchical word embeddings.

\noindent \textbf{Discussion on Hyperbolic Embedding Literature.} Similar regularization approaches have been used in the hyperbolic embedding literature to prevent numerical issues when optimizing hyperbolic embeddings \cite{liu2019hyperbolic,nickel2018learning}. In contrast, our work is focused on the \textit{hyperbolic neural networks} for image classification and its unique \textit{vanishing gradient issue}, which is drastically different from \cite{liu2019hyperbolic,nickel2018learning} in terms of model architecture and the focused problem. In hyperbolic neural networks, the gradients are backpropagated through the hyperbolic layers to the Euclidean layers which causes the gradients to vanish. The vanishing gradient issue will not occur in \cite{liu2019hyperbolic,nickel2018learning} since Euclidean layers are not adopted. 

Lorentz model is used recently to overcome the numerical issues of Poincar\'e ball model for learning word embeddings \cite{nickel2018learning}. However, it is most effective only in low dimensions \cite{liu2019hyperbolic}. For image datasets of ImageNet-scale, hyperbolic neural networks with high-dimensional embeddings are necessary for enough model capacity.

\begin{figure}[!t]
   \centering
\begin{tabular}{c}
 \includegraphics[width=0.35\textwidth]{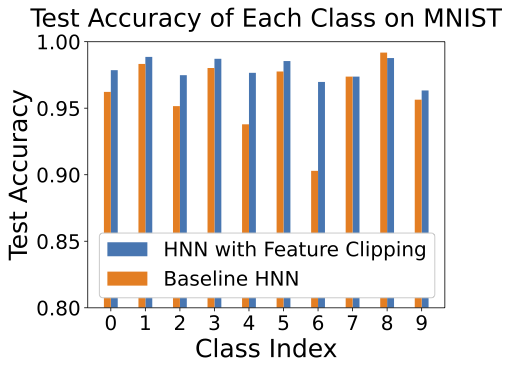} 
 \\
 a) Accuracy of each class on MNIST. \\
\includegraphics[width=0.31\textwidth]{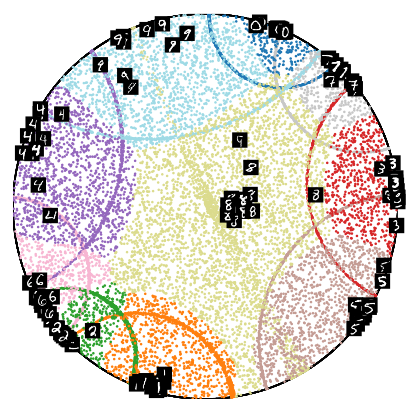} \\ 
 b) Baseline HNNs \\
\includegraphics[width=0.31\textwidth]{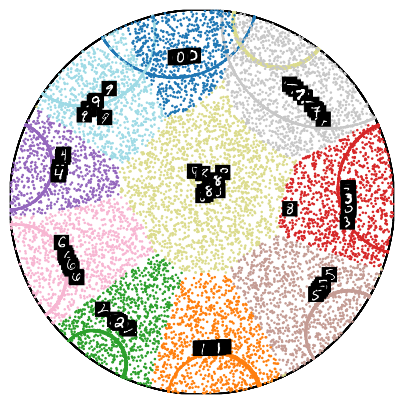} \\
 c) Clipped HNNs \\
\end{tabular}
    \caption{Clipped HNNs greatly outperform baseline HNNs. a) The per class test accuracy of baseline HNNs and Clipped HNNs. b) The Poincar\'e decision hyperplanes and the hyperbolic embeddings of sampled test images of baseline HNNs. c) The Poincar\'e decision hyperplanes and the hyperbolic embeddings of sampled test images of clipped HNNs. Clipped HNNs learn more discriminative feature in hyperbolic space. The per class accuracy indicates that baseline HNNs learn biased feature space which hurts the performance on certain classes. }
     \label{fig:ball_comparison}
\end{figure}

\section{Experimental Results}
\label{sec: exp_setup}

We conduct four types of experiments: standard balanced classification tasks, few-shot learning tasks, adversarial robustness and out-of-distribution detection. The results show that clipped HNNs are on par with ENNs on standard recognition datasets while demonstrating better performance in terms of few-shot classification, adversarial robustness and out-of-distribution detection.

\noindent \textbf{Datasets.} We consider four commonly used image classification datasets: MNIST \cite{lecun1998mnist}, CIFAR10 \cite{krizhevsky2009learning}, CIFAR100 \cite{krizhevsky2009learning} and ImageNet \cite{deng2009imagenet}. See details in the Supplementary. To our best knowledge, this paper is the first attempt to extensively evaluate hyperbolic neural networks on the standard image classification datasets for supervised classification.
 
\noindent \textbf{Baselines and Networks.}  We compare the performance of HNNs training with/without the proposed feature clipping method \cite{ganea2018hyperbolic,khrulkov2020hyperbolic} and their Euclidean counterparts. For MNIST, we use a LeNet-like convolutional neural network \cite{lecun1998gradient} which has two convolutional layers with max pooling layers in between and three fully connected layers. For CIFAR10 and CIFAR100, we use WideResNet \cite{zagoruyko2016wide}. For ImageNet, we use a standard ResNet18 \cite{he2016deep}.

\noindent \textbf{Training Setups.} For training ENNs, we use SGD with momentum. For training HNNs, the Euclidean parameters of HNNs are trained using SGD, and the hyperbolic parameters of HNNs are optimized using stochastic Riemann gradient descent \cite{bonnabel2013stochastic}, just like the previous method. For training networks on MNIST, we train the network for 10 epochs with a learning rate of 0.1. The batch size is 64. For training networks on CIFAR10 and CIFAR100, we train the network for 100 epochs with an initial learning rate of 0.1 and use cosine learning rate scheduler \cite{loshchilov2016sgdr}. The batch size is 128. For training networks on ImageNet, we train the network for 100 epochs with an initial learning rate of 0.1 and the learning rate decays by 10 every 30 epochs. The batch size is 256. We find the HNNs are robust to the choice of the hyperparameter $r$, thus we fix $r$ to be 1.0 in all the experiments. For more discussions and results on the effect of $r$, please see the Supplementary. For baseline HNNs, we use a clipping value of 15 similar to \cite{liu2019hyperbolic,nickel2018learning} to address the numerical issue. The experiments on MNIST, CIFAR10 and CIFAR100 are repeated for 5 times and we report both average accuracy and standard deviation. All the experiments are done on 8 NVIDIA TITAN RTX GPUs. 


\noindent \textbf{Results on Standard Benchmarks.} Table \ref{table: on_all_datasets} shows the results of different networks on the considered benchmarks. On MNIST, we can observe that the accuracy of the improved clipped HNNs is about 5\% higher than the baseline HNNs and match the performance of ENNs. On CIFAR10, CIFAR100 and ImageNet, the improved HNNs achieve 6\%, 3\% and 3\% improvement over baseline HNNs. The results show that HNNs can perform well even on datasets which lack explicit hierarchical structure. 

Figure \ref{fig:ball_comparison} shows the Poincar\'e hyperplanes of all the classes and the hyperbolic embeddings of 1000 sampled test images extracted by the baseline HNNs and clipped HNNs. Note that the Poincar\'e hyperplanes consist of arcs of Euclidean circles that are orthogonal to the boundary of the ball. We also color the points in the ball based on the classification results. It can be observed that by regularizing the magnitude of the hyperbolic embedding, all the embeddings locate in a restricted region of the whole Poincar\'e ball and the network learns more regular and discriminative features in hyperbolic space.


\begin{table}[!t]
\renewcommand{\arraystretch}{1}
\setlength{\tabcolsep}{3pt}
\center
\small
\begin{tabular}{ c |l| l| l} 
\toprule
\multicolumn{4}{c}{ \bf Standard Classification}  \\ \midrule
 
\textbf{Task}  & Euclidean \cite{he2016deep} &  Hyperbolic \cite{ganea2018hyperbolic}  &  C-Hyperbolic \\
 \midrule
 MNIST &  99.12$\pm$0.34 &   94.42$\pm$0.29 &  99.08$\pm$0.31  \\
  \midrule
  CIFAR10 & 94.81$\pm$0.42  &  88.82$\pm$0.51  &  94.76$\pm$0.44 \\
   \midrule
  CIFAR100 & 76.24$\pm$0.35   &  72.26$\pm$0.41   & 75.88$\pm$0.38  \\
   \midrule
  ImageNet &  69.82 &  65.74   & 68.45   \\
  \midrule\midrule
\multicolumn{4}{c}{\bf Few-Shot Classification on CUB Dataset}  \\
 \midrule
  1-Shot 5-Way & 51.31$\pm$0.91  & 61.18$\pm$0.24  &  64.66$\pm$0.24 \\
   \midrule
 5-Shot 5-Way & 70.77$\pm$0.69   & 79.51$\pm$0.16  &  81.76$\pm$0.15 \\
   \midrule\midrule
 \multicolumn{4}{c}{\bf Few-Shot Classification on MiniImageNet Dataset}  \\
  \midrule
  1-Shot 5-Way & 49.42$\pm$0.78 & 51.88$\pm$0.20  &  53.01$\pm$0.22 \\
   \midrule
  5-Shot 5-Way & 51.88$\pm$0.20  &    72.63$\pm$0.16 &   72.66$\pm$0.15 \\

    \bottomrule
\end{tabular}
\vspace{5pt}
\caption{Clipped HNN approaches ENN on standard classification benchmarks. Clipped hyperbolic ProtoNet (C-Hyperbolic) greatly outperforms standard hyperbolic ProtoNet (Hyperbolic) and Euclidean ProtoNet (Euclidean) on few-shot learning tasks.}
\label{table: on_all_datasets}
\end{table}

\noindent \textbf{Few-Shot Learning.} We show that the proposed feature clipping can also improve the performance of Hyperbolic ProtoNet \cite{khrulkov2020hyperbolic} for few-shot learning. Different from the standard ProtoNet \cite{snell2017prototypical} which computes the prototype of each class in Euclidean space, Hyperbolic ProtoNet computes the class prototype in hyperbolic space using hyperbolic averaging. Hyperbolic features are shown to be more effective than Euclidean features for few-shot learning \cite{khrulkov2020hyperbolic}.

We follow the experimental settings in \cite{khrulkov2020hyperbolic} and conduct experiments on CUB \cite{WelinderEtal2010} and miniImageNet  dataset \cite{russakovsky2015imagenet}. We consider 1-shot 5-way and 5-shot 5-way tasks as in \cite{khrulkov2020hyperbolic}. The evaluation is repeated for 10000 times and we report the average performance and the 95\% confidence interval. Table \ref{table: on_all_datasets} shows that the proposed feature clipping further improves the accuracy of Hyperbolic ProtoNet for few-shot classification by as much as 3\%.

\begin{figure*}[ht]
     \centering
     \begin{subfigure}[b]{0.31\textwidth}
         \centering
         \includegraphics[width=1.0\textwidth]{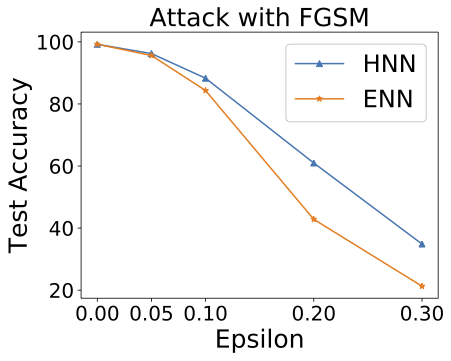}
         \caption{\scriptsize On MNIST}
     \end{subfigure}
     \hfill
     \begin{subfigure}[b]{0.31\textwidth}
         \centering
         \includegraphics[width=1.0\textwidth]{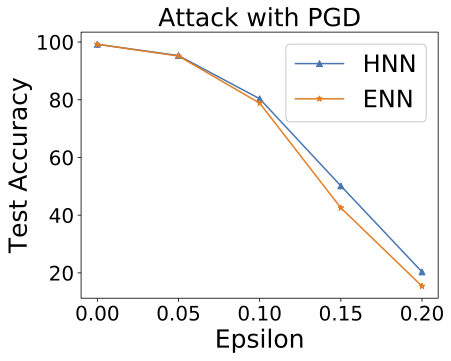}
         \caption{\scriptsize On MNIST}

     \end{subfigure}
          \hfill
     \begin{subfigure}[b]{0.31\textwidth}
         \centering
         \includegraphics[width=1.0\textwidth]{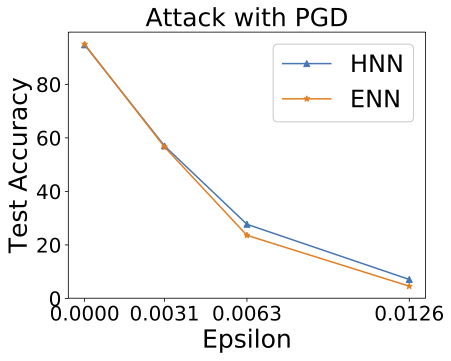}
         \caption{\scriptsize On CIFAR10}
     \end{subfigure}
      
     \caption{ Clipped HNNs are more robust to ENNs against adversarial attacks. We show the results of \textbf{adversarial robustness} of clipped HNNs and ENNs to different attack methods and perturbations. Clipped HNNs are consistently better than ENNs. }
     
     \label{fig:adversarial_figs}
\end{figure*}

\begin{table}[!htt]
\renewcommand{\arraystretch}{1.0}
\centering
\caption{\bf Clipped HNNs consistently outperform ENNs (shaded in gray) on out-of-distribution (OOD) detection with {\it softmax scores} when trained on CIFAR10 and tested on OOD datasets.}
\setlength{\tabcolsep}{4pt}
\begin{tabular}{@{}c| l | l | l}
\toprule
 \textbf{OOD Dataset}
    &$\mathbf{FPR95} \downarrow$  & $\mathbf{AUROC}\uparrow$  & $\mathbf{AUPR}\uparrow$   \\
\midrule

\multirow{2}{*}{ISUN}&
\cellcolor[gray]{0.9}46.30$\pm$0.78&
\cellcolor[gray]{0.9}91.50$\pm$0.16&
\cellcolor[gray]{0.9}98.16$\pm$0.05\\ 
& 
45.28$\pm$0.65&
91.61$\pm$0.21&  
98.09$\pm$0.06\\
\midrule
\multirow{2}{*}{Place365 } & 
\cellcolor[gray]{0.9}51.09$\pm$0.92&
\cellcolor[gray]{0.9}87.56$\pm$0.37&
\cellcolor[gray]{0.9}96.76$\pm$0.15\\
&54.77$\pm$0.76& 
86.82$\pm$0.41&
96.17$\pm$0.20\\
\midrule
\multirow{2}{*}{Texture  }& 
\cellcolor[gray]{0.9}65.04$\pm$0.91&
\cellcolor[gray]{0.9}82.80$\pm$0.35&
\cellcolor[gray]{0.9}94.59$\pm$0.20\\
& 
47.12$\pm$0.62&
89.91$\pm$0.20&
97.39$\pm$0.09\\
\midrule
\multirow{2}{*}{SVHN  }&
\cellcolor[gray]{0.9}71.66$\pm$0.84&
\cellcolor[gray]{0.9}86.58$\pm$0.21&
\cellcolor[gray]{0.9}97.06$\pm$0.06\\
& 
49.89$\pm$1.03& 
91.34$\pm$0.22& 
98.13$\pm$ 0.06\\
\midrule
\multirow{2}{*}{LSUN-Crop}& 
\cellcolor[gray]{0.9}22.22$\pm$0.78&
\cellcolor[gray]{0.9}96.05$\pm$0.10&
\cellcolor[gray]{0.9}99.16$\pm$0.03\\
 & 
23.87$\pm$0.73& 
95.65$\pm$0.22& 
98.98$\pm$0.07\\
\midrule
\multirow{2}{*}{LSUN-Resize}&
\cellcolor[gray]{0.9}41.06$\pm$1.07&
\cellcolor[gray]{0.9}92.67$\pm$0.16& 
\cellcolor[gray]{0.9}98.42$\pm$0.04\\
 & 
 41.49$\pm$1.24&
 92.97$\pm$0.24& 
 98.46 $\pm$0.07\\
 \midrule
\multirow{2}{*}{\textbf{Mean} }&
\cellcolor[gray]{0.9}49.56&
\cellcolor[gray]{0.9}89.53& 
\cellcolor[gray]{0.9}97.36\\
&
\textbf{43.74}&
\textbf{91.38}&
\textbf{97.87}\\
 \bottomrule
\end{tabular}
\label{tab:cifar_10_ood_softmax}
\end{table}

\begin{table}[!ht]
\renewcommand{\arraystretch}{1}
\centering
\caption{\bf Clipped HNNs consistently outperform ENNs (shaded in gray) on out-of-distribution (OOD) detection with {\it softmax scores} when trained on CIFAR100 and tested on OOD datasets.  On average, they are on par with ENNs by AUPR and far better by PRR95 and AUROC.}

\setlength{\tabcolsep}{4pt}
\begin{tabular}{@{}c | l | l | l}
\toprule
 \textbf{OOD Dataset}
    &$\mathbf{FPR95} \downarrow$  & $\mathbf{AUROC}\uparrow$  & $\mathbf{AUPR}\uparrow$   \\
\midrule
\multirow{2}{*}{ISUN} &
\cellcolor[gray]{0.9}74.07$\pm$0.87&
\cellcolor[gray]{0.9}82.51$\pm$0.39&
\cellcolor[gray]{0.9}95.83$\pm$0.11\\
& 
68.37$\pm$0.90&
81.31$\pm$0.43&
94.96$\pm$0.20\\
\midrule
\multirow{2}{*}{Place365} &
\cellcolor[gray]{0.9}81.01$\pm$1.07&
\cellcolor[gray]{0.9}76.90$\pm$0.45&
\cellcolor[gray]{0.9}94.02$\pm$0.15\\
& 
79.66$\pm$0.69&
76.94$\pm$0.28&
93.91$\pm$0.18\\
\midrule 
\multirow{2}{*}{Texture  }  &
\cellcolor[gray]{0.9}83.67$\pm$0.68&
\cellcolor[gray]{0.9}77.52$\pm$0.32& 
\cellcolor[gray]{0.9}94.47$\pm$0.10\\
&
64.91$\pm$0.80&
83.26$\pm$0.25& 
95.77$\pm$0.08\\
\midrule
\multirow{2}{*}{SVHN  } &
\cellcolor[gray]{0.9}84.56$\pm$0.78&
\cellcolor[gray]{0.9}84.32$\pm$0.22&
\cellcolor[gray]{0.9}96.69$\pm$0.07\\
&
53.11$\pm$1.04&
89.53$\pm$0.26&
97.71$\pm$0.07\\
\midrule
\multirow{2}{*}{LSUN-Crop   }  &
\cellcolor[gray]{0.9}43.46$\pm$0.79&
\cellcolor[gray]{0.9}93.09$\pm$0.23& 
\cellcolor[gray]{0.9}98.58$\pm$0.05\\
&
51.08$\pm$1.17&
87.21$\pm$0.39&
96.83$\pm$0.13\\
\midrule
\multirow{2}{*}{LSUN-Resize  } &
\cellcolor[gray]{0.9}71.50$\pm$0.73&
\cellcolor[gray]{0.9}82.12$\pm$0.40&
\cellcolor[gray]{0.9}95.69$\pm$0.13\\
&
63.86$\pm$1.10&
82.36$\pm$0.42&
95.16$\pm$0.13\\
\midrule
\multirow{2}{*}{\textbf{Mean} }&
\cellcolor[gray]{0.9}73.05&
\cellcolor[gray]{0.9}82.74&
\cellcolor[gray]{0.9}\textbf{95.88}\\
& 
\textbf{63.50}&
\textbf{83.43}& 
95.72\\
 \bottomrule
\end{tabular}
\label{tab:cifar_100_ood_softmax}
\end{table}

\noindent \textbf{Adversarial Robustness.} We show that clipped HNNs are more robust to adversarial attacks including FGSM \cite{goodfellow2014explaining} and PGD \cite{madry2017towards} than ENNs. For attacking networks trained on MNIST using FGSM, we consider the perturbation $\epsilon = 0.05, 0.1, 0.2, 0.3$. For attacking networks trained on MNIST using PGD, we consider the perturbation $\epsilon = 0.05, 0.1, 0.15, 0.2$. The number of steps is 40. For attacking networks trained on CIFAR10 using PGD, we consider the perturbation $\epsilon = 0.8/255, 1.6/255, 3.2/255$. The number of steps is 7. 

From Figure \ref{fig:adversarial_figs} we can see that across all the cases clipped HNNs show more robustness than ENNs to adversarial attacks. For more discussions and results using vanilla HNNs, please see the Supplementary.

\noindent \textbf{Out-of-Distribution Detection.} We conduct experiments to show that clipped HNNs have stronger out-of-distribution detection capability than ENNs. Out-of-distribution detection aims at determining whether or not a given input is from the same distribution as the training data. We follow the experimental settings in \cite{liu2020energy}. The in-distribution datasets are CIFAR10 and CIFAR100. The out-of-distribution datasets are ISUN \cite{xu2015turkergaze}, Place365 \cite{zhou2017places}, Texture \cite{cimpoi2014describing}, SVHN \cite{netzer2011reading}, LSUN-Crop \cite{yu2015lsun} and LSUN-Resize \cite{yu2015lsun}. For detecting out-of-distribution data, we use both softmax score and energy score as described in \cite{liu2020energy}. For metrics, we consider FPR95, AUROC and AUPR \cite{liu2020energy}. Table \ref{tab:cifar_10_ood_softmax} and \ref{tab:cifar_100_ood_softmax} show the results of using softmax score on CIFAR10 and CIFAR100 respectively. We can see that HNNs and ENNs achieve similar AUPR, however HNNs achieve much better performance in terms of FPR95 and AUROC. In particular, HNNs reduce FPR95 by 5.82\% and 9.55\% on CIFAR10 and CIFAR100 respectively. For results using energy score and vanilla HNNs, please see the Supplementary.

\begin{figure}[!ht]
   \centering
\begin{tabular}{cc}
\includegraphics[width=0.22\textwidth]{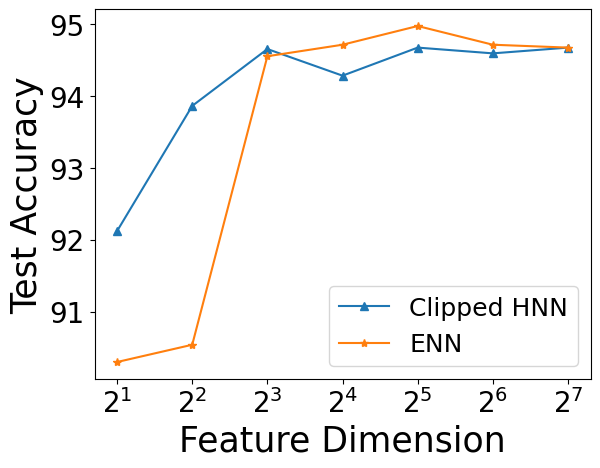}& 
\includegraphics[width=0.22\textwidth]{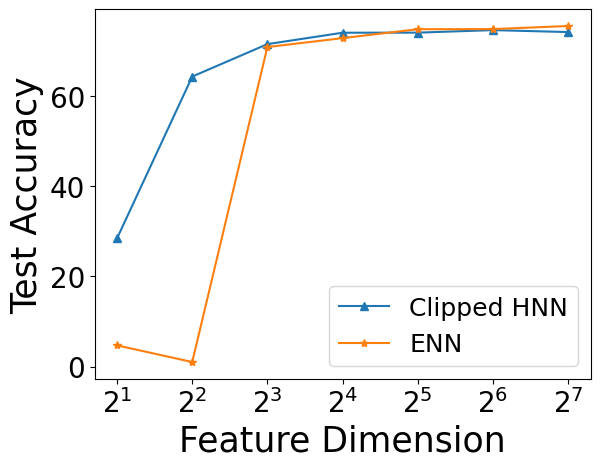}
\end{tabular}·
    \caption{Clipped HNNs are better than ENNs when the feature dimension is low. \textbf{Left}: Test accuracy on CIFAR10. \textbf{Right}: Test accuracy on CIFAR100. We change the feature dimension from 2 to 128. }
     \label{fig:ablation_on_dimension}
\end{figure}

\noindent \textbf{The Effect of Feature Dimension.} Figure \ref{fig:ablation_on_dimension} shows the change of test accuracy as we vary the feature dimension on CIFAR10 and CIFAR100. Clipped HNNs are much better than ENNs when the feature dimension is low. One possible reason is that when the dimension is low in the Euclidean case, the data are hard to be linearly separated. However in hyperbolic space, since the Poincar\'e hyperplanes are ``curved'', the data are more likely to be linearly separated even in two dimensions. 

\section{Summary}

We propose a simple yet effective solution called \emph{Feature Clipping} to address the vanishing gradient problem in training HNNs. We conduct extensive experiments on commonly used image dataset benchmarks. To the best of our knowledge, this is the first time that HNNs can be applied to image datasets of ImageNet-scale. Clipped HNNs show significant improvement over baseline HNNs and match the performance of ENNs. The proposed feature clipping also improves the performance of HNNs for few-shot learning. Further experimental studies reveal that clipped HNNs are more robust to adversarial attacks such as PGD and FGSM. Clipped HNNs also show stronger out-of-distribution detection capability than ENNs. When the feature dimension is low, clipped HNNs even outperform ENNs.

{\bf Acknowledgements.} 
This work was supported, in part, by Berkeley Deep Drive and the National Science Foundation under Grant No. 2131111. Any opinions, findings, and conclusions or recommendations expressed here are those of the authors and do not necessarily reflect the views of the National Science Foundation.

\section{Organization of the supplementary}

\subsection{Background and datasets}
\begin{enumerate}
    \item In Section \ref{sec: append_gyro}, we introduce the background on Gyrovector space.
        \item In Section \ref{sec:datasets}, we give the detailed statistics of the datasets.
\end{enumerate}

\subsection{Effect of the gradient update}
 In Section \ref{sec: effect_of_update}, we derive the effect of the gradient update of the Euclidean parameters on the hyperbolic embeddings. 
\subsection{Effect of the clipped value $r$} 
In Section \ref{sec:ablation_on_r}, we show the effect of different choices of $r$. \emph{To conclude, there is a sweet spot in terms of choosing $r$ which is neither too large (causing vanishing gradient problem) nor too small (not enough capacity). The performance of clipped HNNs is also robust to the choice of the hyperparameter $r$ if it is around the sweet spot.}

\subsection{Adversarial robustness}
In Section \ref{sec:more_on_adv}, we show more results on adversarial robustness. \emph{Clipped HNNs show more robustness to ENNs and greatly improve the robustness of vanilla HNNs.}

\subsection{More on out-of-distribution detection}

In Section \ref{sec:more_on_ood}, we show more results on out-of-distribution (OOD) detection using ENNs, vanilla HNNs and clipped HNNs. \emph{Clipped HNNs show stronger OOD detection capability compared with ENNs and greatly improve the OOD detection capability of vanilla HNNs.} 

\subsection{Using softmax with temperature scaling as a workaround}
In Section \ref{sec:temperature} we show that using softmax with temperature scaling as a workaround for addressing the vanishing gradient problem. \emph{When feature dimension is high, softmax with temperature scaling severely underperforms the proposed feature clipping. The results again confirm the effectiveness of the proposed approach.}

\subsection{Clipped hyperbolic space is still hyperbolic}

In Section \ref{sec:still_hyperbolic}, we show that clipped hyperbolic space is still hyperbolic. \emph{Using clipped hyperbolic space for learning word embeddings is better than using the unclipped version.}

\subsection{With norm regularization term}

In Section \ref{sec:with_regularization}, we show the results using norm regularization during training. \emph{The proposed feature clipping outperforms using the norm regularization term.}

\subsection{More discussions on Lorentz model}

In Section \ref{sec:Lorentz}, we give more discussions on Lorentz model and why we focus on hyperbolic neural networks based on Poincar\'e ball model.

\newpage
\section{Gyrovector space}
\label{sec: append_gyro}

We give more details on gyrovector space, for a more systematic treatment, please refer to \cite{ungar2005analytic,ungar2008gyrovector,ungar2001hyperbolic}.

Gyrovector space provides a way to operate in hyperbolic space with vector algebra. Gyrovector space to hyperbolic geometry is similar to standard vector space to Euclidean geometry. The geometric objects in gyrovector space are called gyroevectors which are equivalent classes of directed gyrosegments. Similar to the vectors in Euclidean space which are added according to parallelogram law, gyrovectors are added according to gyroarallelogram law. Technically, gyrovector spaces are gyrocommutative gyrogroups of gyrovectors that admit scalar multiplications. 

We start from the introduction of gyrogroups which give rise to gyrovector spaces. 

\begin{definition}[Gyrogroups]

A groupoid $(G, \oplus)$ is a gyrogroup if it satisfies the follow axioms,
\begin{enumerate}
    \item There exist one element $0 \in G$ satisfies $0 \oplus a = a$ for all $a \in G$.
    \item For each $a \in G$, there exist an element $\ominus a \in G$ which satisfies $\ominus a \oplus a = 0$
    \item For every $a, b, c \in G$, there exist a unique element gry$[a,b]c \in G$ such that $\oplus$ satisfies the left gyroassociative law $a\oplus(b\oplus c) = (a\oplus b)\oplus$gry$[a,b]c$. 
    \item The map gry$[a,b]c$: $G \rightarrow G$ given by $c  \mapsto$ gry$[a,b]c$ is an automorphism of the groupoid $(G, \oplus)$: gyr$[a,b] \in Aut(G, \oplus)$. The automorphism gyr$[a,b]$ of $G$ is called the gyroautomorphism of $G$ generated by $a, b \in G$.
    \item The operation gry: $G \times G \rightarrow Aut(G, \oplus)$ is called gyrator of $G$. The gyroautomorphism gyr$[a,b]$ generated by any $a, b \in G$ has the left loop property: gyr$[a,b]$ = gyr$[a \oplus b, b]$.
\end{enumerate}
\end{definition}

In particular, M\"{o}bius complex disk groupoid $(\mathbb{D}, \oplus_{M})$ is a gyrocommunicative gyrogroup, where $\mathbb{D} = \{ z \in \mathbb{C}: |z| < 1 \}$ and $\oplus_{M}$ is the M\"{o}bius addition. The same applies to the $s$-ball $\mathbb{V}_s$ which is defined as,
\begin{equation}
    \mathbb{V}_s = \{\mathbf{v} \in \mathbb{V}: \lVert \mathbf{v} \rVert < s\}
\end{equation}

Gyrocommutative gyrogroups which admit scalar multiplication $\oplus$ become gyrovector space $(G, \oplus, \otimes)$. M\"{o}bius gyrogroups $(\mathbb{V}, \oplus_{M})$ admit scalar multiplication $\oplus_M$ become M\"{o}bius gyrovector space $(\mathbb{V}, \oplus_M, \otimes_M)$. 
\begin{definition} [M\"{o}bius Scalar Multiplication]
Let $(\mathbb{V}_s, \oplus_{M})$ be a M\"{o}bius gyrogroup, the M\"{o}bius scalar multiplication $\otimes_M$ is defined as,
\begin{equation}
    r \otimes_M \mathbf{v} = s \frac{(1+\frac{\lVert \mathbf{v} \rVert}{s})^r - (1 - \frac{\lVert \mathbf{v} \rVert}{s})^r}{(1+\frac{\lVert \mathbf{v} \rVert}{s})^r + (1 - \frac{\lVert \mathbf{v} \rVert}{s})^r} \frac{\mathbf{v}}{\lVert \mathbf{v} \rVert}
\end{equation}
where $r \in \mathbb{R}$ and $\mathbf{v} \in \mathbb{V}_s$, $\mathbf{v} \neq \mathbf{0}$.
\end{definition}

\begin{definition}[Gyrolines]
Let $\mathbf{a}, \mathbf{b}$ be two distinct points in the gyrovector space $(G, \oplus, \otimes)$. The gyroline in $G$ which passes through $\mathbf{a}, \mathbf{b}$ is the set of points:
\begin{equation}
    L = \mathbf{a}\oplus(\ominus\mathbf{a}\oplus\mathbf{b}) \otimes t
\end{equation}
where $t \in \mathbb{R}$.
\end{definition}
\begin{figure*}[!ht]
   \centering
\begin{tabular}{ccc}
\includegraphics[width=0.3\textwidth]{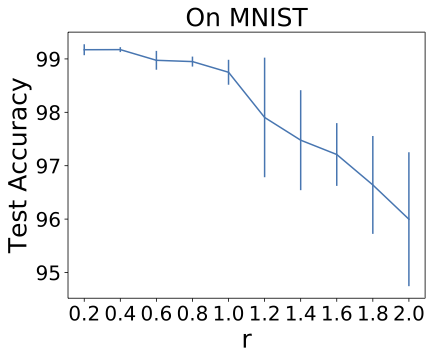}& \includegraphics[width=0.3\textwidth]{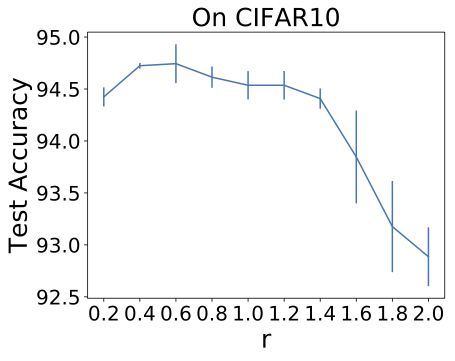} &
\includegraphics[width=0.3\textwidth]{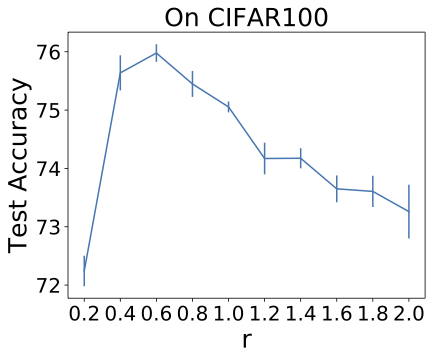}
\end{tabular}
    \caption{We show the change of the test accuracy as we vary the hyperparameter $r$. A large $r$ leads to \emph{vanishing gradient problem} and a small $r$ causes insufficient capacity. Both lead to a drop in test accuracy.   }
     \label{fig:ablation_on_r}
\end{figure*}

It can be proven that gyrolines in a M\"{o}bius gyrovector space coincide with the geodesics of the Poincar\'e ball model of hyperbolic geometry.

With the aid of operations in gyrovector spaces, we can define important properties of the Poincar\'e ball model in closed-form expressions. 

\begin{definition}[Exponential Map and Logarithmic Map]
As shown in \cite{ganea2018hyperbolic}, the exponential map $\exp^c_{\mathbf{x}}: T_\mathbf{x}\mathbb{B}^n_c \rightarrow \mathbb{B}^n_c$ is defined as,
\begin{equation}
\scriptsize
    \exp^c_{\mathbf{x}}(\mathbf{v}) = \mathbf{x} \oplus_c (\tanh (\frac{\sqrt{c}\lambda_{\mathbf{x}}^c\lVert \mathbf{v} \rVert}{2})\frac{\mathbf{v}}{\sqrt{c}\lVert \mathbf{v} \rVert}), \quad\quad \forall \mathbf{x} \in \mathbb{B}_c^n, \mathbf{v} \in T_\mathbf{x}\mathbb{B}^n_c
\end{equation}
\end{definition}

The logarithmic map $\log_{\mathbf{x}}^c: \mathbb{B}^n_c \rightarrow T_\mathbf{x}\mathbb{B}^n_c$ is defined as,
\begin{equation}
\scriptsize
    \log_{\mathbf{x}}^c = \frac{2}{\sqrt{c}\lambda^c_{\mathbf{x}}}\tanh^{-1}(\sqrt{c}\lVert \ominus_c \mathbf{x}\oplus_c \mathbf{y} \rVert) \frac{\ominus_c \mathbf{x}\oplus_c \mathbf{y}}{\lVert \ominus_c \mathbf{x}\oplus_c \mathbf{y} \rVert }, 
    \quad\quad \mathbf{x}, \mathbf{y} \in \mathbb{B}^n_c
\end{equation}

The distance between two points in the Poincar\'e ball can be defined as,
\begin{definition}(Poincar\'e Distance between Two Points)
\begin{equation}
    d_c(\mathbf{x}, \mathbf{y}) = \frac{2}{\sqrt{c}}\tanh^{-1} (\sqrt{c}\lVert  \ominus_c \mathbf{x}\oplus_c \mathbf{y} \rVert)
\end{equation}

\end{definition}

\begin{table}[!ht]
\renewcommand{\arraystretch}{1.15}
\centering
\caption{\textbf{The results of out-of-distribution detection} on \textcolor{red}{CIFAR10} with \textcolor{red}{softmax score}. The results of ENNs are shaded in dark gray. The results of vanilla HNNs are shaded in light gray. Clipped HNNs achieve higher average performance across all the datasets and greatly improve the OOD detection capability of vanilla HNNs.}
\scalebox{0.8}{
\begin{tabular}{c | c | c | c}
\toprule
\textbf{OOD Dataset}
    &$\mathbf{FPR95} \downarrow$  & $\mathbf{AUROC}\uparrow$  & $\mathbf{AUPR}\uparrow$   \\
\midrule
\midrule
\multirow{3}{*}{ISUN}&\cellcolor[gray]{0.8} 46.30 $\pm$ 0.78&\cellcolor[gray]{0.8}  91.50 $\pm$  0.16&\cellcolor[gray]{0.8} 98.16 $\pm$ 0.05  \\
&\cellcolor[gray]{0.9}  98.37 $\pm$ 0.20 &\cellcolor[gray]{0.9} 29.72 $\pm$ 0.58 & \cellcolor[gray]{0.9} 74.54 $\pm$ 0.19 \\
& 45.28 $\pm$ 0.65 &91.61 $\pm$ 0.21 &  98.09 $\pm$ 0.06 \\
\midrule
\multirow{3}{*}{Place365 } & \cellcolor[gray]{0.8}  51.09  $\pm$ 0.92 &\cellcolor[gray]{0.8} 87.56 $\pm$ 0.37&\cellcolor[gray]{0.8}  96.76 $\pm$ 0.15 \\
& \cellcolor[gray]{0.9}  96.10 $\pm$ 0.32 &\cellcolor[gray]{0.9}  44.82 $\pm$ 0.65 & \cellcolor[gray]{0.9} 80.67 $\pm$ 0.34   \\

 &54.77 $\pm$ 0.76& 86.82 $\pm$ 0.41&96.17 $\pm$ 0.20 \\
        \midrule
\multirow{3}{*}{Texture  }& \cellcolor[gray]{0.8} 65.04 $\pm$ 0.91&\cellcolor[gray]{0.8} 82.80 $\pm$ 0.35 &\cellcolor[gray]{0.8} 94.59 $\pm$0.20 \\
&\cellcolor[gray]{0.9} 97.62 $\pm$ 0.15 &\cellcolor[gray]{0.9}33.87 $\pm$ 0.40 & \cellcolor[gray]{0.9} 74.52 $\pm$ 0.17 \\
 & 47.12 $\pm$ 0.62&89.91 $\pm$ 0.20&97.39 $\pm$ 0.09 \\
    \midrule
\multirow{3}{*}{SVHN  }&\cellcolor[gray]{0.8}  71.66 $\pm$ 0.84&\cellcolor[gray]{0.8} 86.58 $\pm$ 0.21&\cellcolor[gray]{0.8} 97.06 $\pm$ 0.06\\
& \cellcolor[gray]{0.9} 91.03 $\pm$ 0.53 &\cellcolor[gray]{0.9}   61.33  $\pm$ 0.53  &\cellcolor[gray]{0.9}  86.39  $\pm$ 0.27\\
& 49.89 $\pm$ 1.03 & 91.34 $\pm$ 0.22& 98.13 $\pm$  0.06  \\
     \midrule
 \multirow{3}{*}{LSUN-Crop   }& \cellcolor[gray]{0.8} 22.22 $\pm$ 0.78&\cellcolor[gray]{0.8} 96.05 $\pm$ 0.10 &\cellcolor[gray]{0.8} 99.16 $\pm$ 0.03 \\
 &\cellcolor[gray]{0.9}  96.18 $\pm$  0.36  &\cellcolor[gray]{0.9}  37.29 $\pm$ 0.63 &\cellcolor[gray]{0.9}  75.94 $\pm$ 0.27  \\
 & 23.87 $\pm$  0.73& 95.65 $\pm$ 0.22& 98.98 $\pm$ 0.07 \\
     \midrule
 \multirow{3}{*}{LSUN-Resize  }& \cellcolor[gray]{0.8} 41.06 $\pm$1.07&\cellcolor[gray]{0.8} 92.67 $\pm$ 0.16 & \cellcolor[gray]{0.8} 98.42 $\pm$ 0.04 \\
 &\cellcolor[gray]{0.9}  99.62 $\pm$ 0.10 &\cellcolor[gray]{0.9}  22.05 $\pm$ 0.32 &\cellcolor[gray]{0.9}  71.88 $\pm$ 0.15  \\
 & 41.49 $\pm$ 1.24 &  92.97 $\pm$ 0.24& 98.46  $\pm$ 0.07  \\
 \midrule
 
  \multirow{3}{*}{\textbf{Mean} }&\cellcolor[gray]{0.8}49.56 &\cellcolor[gray]{0.8}89.53 & \cellcolor[gray]{0.8} 97.36   \\
  &\cellcolor[gray]{0.9}  96.49 & \cellcolor[gray]{0.9} 38.18 &\cellcolor[gray]{0.9}   77.32  \\
  &\textbf{43.74} &\textbf{91.38}&\textbf{97.87}\\
  
 \bottomrule
\end{tabular}}
\label{tab:cifar_10_ood_softmax_s}
\end{table}

\begin{table}[!ht]
\renewcommand{\arraystretch}{1.15}
\centering
\caption{\textbf{The results of out-of-distribution detection} on \textcolor{red}{CIFAR100} with \textcolor{red}{softmax score}. The results of ENNs are shaded in dark gray. The results of vanilla HNNs are shaded in light gray. Clipped HNNs achieve comparable performance to ENNs in terms of AUPR and higher average performance in terms of PRR95 and AUROC.}
\scalebox{0.8}{
\begin{tabular}{c | c | c | c}
\toprule
\textbf{OOD Dataset}
    &$\mathbf{FPR95} \downarrow$  & $\mathbf{AUROC}\uparrow$  & $\mathbf{AUPR}\uparrow$   \\
\midrule
\midrule
\multirow{3}{*}{ISUN}  &\cellcolor[gray]{0.8} 74.07 $\pm$ 0.87&\cellcolor[gray]{0.8} 82.51 $\pm$ 0.39 &\cellcolor[gray]{0.8} 95.83 $\pm$ 0.11 \\
& \cellcolor[gray]{0.9 } 80.97 $\pm$ 0.65 &\cellcolor[gray]{0.9} 69.24  $\pm$ 0.52 &\cellcolor[gray]{0.9} 89.98 $\pm$  0.22 \\
& 68.37 $\pm$ 0.90 & 81.31 $\pm$ 0.43& 94.96 $\pm$ 0.20\\
\midrule
\multirow{3}{*}{Place365 } &\cellcolor[gray]{0.8} 81.01 $\pm$ 1.07&\cellcolor[gray]{0.8} 76.90 $\pm$ 0.45&\cellcolor[gray]{0.8} 94.02 $\pm$ 0.15 \\
&\cellcolor[gray]{0.9}82.75 $\pm$ 0.66 &\cellcolor[gray]{0.9} 71.97 $\pm$ 0.50 &\cellcolor[gray]{0.9} 92.27 $\pm$ 0.22\\
& 79.66 $\pm$ 0.69 &76.94 $\pm$ 0.28 &93.91 $\pm$ 0.18 \\
        \midrule 
\multirow{3}{*}{Texture  }  &\cellcolor[gray]{0.8} 83.67 $\pm$ 0.68&\cellcolor[gray]{0.8} 77.52 $\pm$ 0.32& \cellcolor[gray]{0.8} 94.47 $\pm$ 0.10 \\
&\cellcolor[gray]{0.9}75.33 $\pm$ 0.92 &\cellcolor[gray]{0.9} 75.14 $\pm$ 0.49 &\cellcolor[gray]{0.9} 92.39 $\pm$ 0.20 \\
&64.91 $\pm$ 0.80  &83.26 $\pm$ 0.25& 95.77 $\pm$ 0.08\\
    \midrule
\multirow{3}{*}{SVHN  } &\cellcolor[gray]{0.8} 84.56 $\pm$ 0.78&\cellcolor[gray]{0.8} 84.32 $\pm$ 0.22 &\cellcolor[gray]{0.8} 96.69 $\pm$ 0.07 \\
&\cellcolor[gray]{0.9} 62.83 $\pm$ 0.70 &\cellcolor[gray]{0.9} 84.29 $\pm$ 0.23 &\cellcolor[gray]{0.9} 96.31 $\pm$  0.06\\
&53.11 $\pm$ 1.04 &89.53 $\pm$ 0.26&97.71 $\pm$ 0.07\\
     \midrule
 \multirow{3}{*}{LSUN-Crop   }  &\cellcolor[gray]{0.8}43.46 $\pm$ 0.79&\cellcolor[gray]{0.8}93.09 $\pm$ 0.23 &\cellcolor[gray]{0.8}98.58 $\pm$ 0.05 \\
 &\cellcolor[gray]{0.9}56.66 $\pm$ 0.67 &\cellcolor[gray]{0.9}89.30 $\pm$ 0.17 &\cellcolor[gray]{0.9}97.71 $\pm$ 0.04 \\
 &51.08 $\pm$ 1.17 &87.21 $\pm$  0.39&96.83 $\pm$ 0.13\\
     \midrule
 \multirow{3}{*}{LSUN-Resize  } &\cellcolor[gray]{0.8}71.50 $\pm$ 0.73&\cellcolor[gray]{0.8}82.12 $\pm$  0.40&\cellcolor[gray]{0.8}95.69 $\pm$ 0.13 \\
 &\cellcolor[gray]{0.9}75.50 $\pm$ 0.81 &\cellcolor[gray]{0.9}73.40 $\pm$ 0.68 &\cellcolor[gray]{0.9} 91.41 $\pm$ 0.28 \\
 &63.86 $\pm$ 1.10 &82.36 $\pm$ 0.42&95.16 $\pm$ 0.13 \\
 \midrule
 
  \multirow{3}{*}{\textbf{Mean} }  &\cellcolor[gray]{0.8}73.05  &\cellcolor[gray]{0.8} 82.74&\cellcolor[gray]{0.8} \textbf{95.88}\\
  &\cellcolor[gray]{0.9} 72.34&\cellcolor[gray]{0.9}77.22 &\cellcolor[gray]{0.9}93.35 \\
  & \textbf{63.50}  &  \textbf{83.43}  & 95.72 \\
  
 \bottomrule
\end{tabular}}
\label{tab:cifar_100_ood_softmax_sup}
\end{table}

\begin{table}[!ht]
\renewcommand{\arraystretch}{1.15}
\centering
\caption{\textbf{The results of out-of-distribution detection} on \textcolor{red}{CIFAR10} with \textcolor{red}{energy score}. The results of ENNs are shaded in dark gray. The results of vanilla HNNs are shaded in light gray.}
\scalebox{0.8}{
\begin{tabular}{c | c | c | c}
\toprule
 \textbf{OOD Dataset}
    &$\mathbf{FPR95} \downarrow$  & $\mathbf{AUROC}\uparrow$  & $\mathbf{AUPR}\uparrow$   \\
\midrule
\midrule
\multirow{3}{*}{ISUN}&\cellcolor[gray]{0.8} 34.19 $\pm$ 0.97  &\cellcolor[gray]{0.8} 93.07 $\pm$ 0.24 &\cellcolor[gray]{0.8} 98.42 $\pm$ 0.07   \\
&\cellcolor[gray]{0.9} 99.31 $\pm$ 0.15 &\cellcolor[gray]{0.9}  28.69 $\pm$  0.35 & \cellcolor[gray]{0.9} 74.14 $\pm$ 0.10\\
& 25.39 $\pm$ 0.32  &95.48 $\pm$ 0.09 &  99.01 $\pm$ 0.04 \\
\midrule
\multirow{3}{*}{Place365 } & \cellcolor[gray]{0.8} 43.34 $\pm$ 1.22  &\cellcolor[gray]{0.8}88.50 $\pm$ 0.48  &\cellcolor[gray]{0.8} 96.76 $\pm$ 0.17   \\
&\cellcolor[gray]{0.9}97.57 $\pm$ 0.37 &\cellcolor[gray]{0.9} 43.96 $\pm$ 0.87 &\cellcolor[gray]{0.9} 80.36 $\pm$ 0.46\\
 & 45.17 $\pm$ 1.19& 89.61 $\pm$ 0.28 & 97.20 $\pm$ 0.14 \\
        \midrule
\multirow{3}{*}{Texture  }& \cellcolor[gray]{0.8} 58.51 $\pm$ 0.77 &\cellcolor[gray]{0.8} 82.98 $\pm$ 0.20 &\cellcolor[gray]{0.8} 94.55 $\pm$ 0.14  \\
&\cellcolor[gray]{0.9} 95.93 $\pm$ 0.29  &\cellcolor[gray]{0.9} 35.02 $\pm$ 0.41 & \cellcolor[gray]{0.9}  74.87 $\pm$ 0.14 \\
 & 49.70 $\pm$ 0.94&90.66 $\pm$ 0.20& 97.98 $\pm$ 0.04\\
    \midrule
\multirow{3}{*}{SVHN  }&\cellcolor[gray]{0.8} 49.04 $\pm$ 1.05 &\cellcolor[gray]{0.8} 91.57 $\pm$ 0.13  &\cellcolor[gray]{0.8} 98.12 $\pm$ 0.05\\
&\cellcolor[gray]{0.9}96.71 $\pm$ 0.37 &\cellcolor[gray]{0.9}59.65 $\pm$ 0.56 &\cellcolor[gray]{0.9}86.16 $\pm$ 0.25 \\
& 57.33 $\pm$ 1.34 & 88.45 $\pm$ 0.20&  97.44 $\pm$ 0.06  \\
     \midrule
 \multirow{3}{*}{LSUN-Crop   }& \cellcolor[gray]{0.8} 9.48 $\pm$ 0.60 &\cellcolor[gray]{0.8}98.21 $\pm$ 0.07 &\cellcolor[gray]{0.8} 99.63 $\pm$ 0.02  \\
 &\cellcolor[gray]{0.9}  98.18 $\pm$  0.27 &\cellcolor[gray]{0.9} 36.34 $\pm$  0.63 &\cellcolor[gray]{0.9} 75.64 $\pm$ 0.26 \\
 &24.78 $\pm$ 0.73   & 95.06 $\pm$ 0.15& 98.92 $\pm$ 0.05 \\
     \midrule
 \multirow{3}{*}{LSUN-Resize  }& \cellcolor[gray]{0.8} 28.28 $\pm$ 0.66 &\cellcolor[gray]{0.8} 94.31 $\pm$ 0.14& \cellcolor[gray]{0.8} 98.72 $\pm$ 0.04 \\
 &\cellcolor[gray]{0.9}99.91 $\pm$ 0.06 &\cellcolor[gray]{0.9}21.34 $\pm$ 0.48 &\cellcolor[gray]{0.9} 71.60 $\pm$ 0.21\\
 &  22.52 $\pm$ 0.67& 96.15 $\pm$ 0.09 & 99.18 $\pm$ 0.02  \\
 \midrule
 
  \multirow{3}{*}{\textbf{Mean} }&\cellcolor[gray]{0.8} \textbf{37.14} &\cellcolor[gray]{0.8}  91.44   &  \cellcolor[gray]{0.8} 97.70  \\
  &\cellcolor[gray]{0.9}97.94 &\cellcolor[gray]{0.9} 37.50&\cellcolor[gray]{0.9} 77.13 \\
  &37.48  &\textbf{92.57} &\textbf{98.29} \\

 \bottomrule
\end{tabular}}
\label{tab:cifar_10_ood_energy}
\end{table}

\begin{table}[!ht]
\renewcommand{\arraystretch}{1.15}
\centering
\caption{\textbf{The results of out-of-distribution detection} on \textcolor{red}{CIFAR100} with \textcolor{red}{energy score}. The results of ENNs are shaded in dark gray. The results of vanilla HNNs are shaded in light gray.}
\scalebox{0.8}{
\begin{tabular}{c | c | c | c}
\toprule
\textbf{OOD Dataset}
    &$\mathbf{FPR95} \downarrow$  & $\mathbf{AUROC}\uparrow$  & $\mathbf{AUPR}\uparrow$   \\
\midrule
\midrule
\multirow{3}{*}{ISUN}  &\cellcolor[gray]{0.8} 74.49 $\pm$ 0.60 &\cellcolor[gray]{0.8}  82.45 $\pm$ 0.33 &\cellcolor[gray]{0.8} 95.84 $\pm$ 0.12 \\
& \cellcolor[gray]{0.9}81.73 $\pm$ 0.54 & \cellcolor[gray]{0.9}70.38 $\pm$ 0.28 & \cellcolor[gray]{0.9}90.76 $\pm$ 0.20 \\
& 68.75 $\pm$ 0.93 & 81.33 $\pm$ 0.31&94.93 $\pm$ 0.16 \\
\midrule
\multirow{3}{*}{Place365 } &\cellcolor[gray]{0.8} 81.20 $\pm$ 0.86 &\cellcolor[gray]{0.8} 77.02 $\pm$ 0.34&\cellcolor[gray]{0.8} 94.13 $\pm$ 0.13 \\
& \cellcolor[gray]{0.9}82.73 $\pm$ 0.98 & \cellcolor[gray]{0.9}74.04 $\pm$  0.55 & \cellcolor[gray]{0.9} 93.20 $\pm$ 0.24\\
& 79.51 $\pm$ 0.69 & 77.23 $\pm$ 0.37& 93.97 $\pm$ 0.17\\
        \midrule 
\multirow{3}{*}{Texture  }  &\cellcolor[gray]{0.8} 83.19 $\pm$ 0.31 &\cellcolor[gray]{0.8} 77.74 $\pm$ 0.35 & \cellcolor[gray]{0.8} 94.54 $\pm$ 0.11  \\
& \cellcolor[gray]{0.9}  72.77 $\pm$ 0.52 & \cellcolor[gray]{0.9} 77.38 $\pm$ 0.39 & \cellcolor[gray]{0.9} 93.38 $\pm$ 0.21 \\
& 65.03 $\pm$ 0.52 &  83.38 $\pm$ 0.29 & 95.85 $\pm$ 0.10\\
    \midrule
\multirow{3}{*}{SVHN  } &\cellcolor[gray]{0.8} 84.12 $\pm$ 0.59 &\cellcolor[gray]{0.8}  84.41 $\pm$ 0.16 &\cellcolor[gray]{0.8} 96.72 $\pm$ 0.04  \\
& \cellcolor[gray]{0.9} 53.37 $\pm$ 0.67 & \cellcolor[gray]{0.9}86.37 $\pm$ 0.30 & \cellcolor[gray]{0.9}96.78 $\pm$ 0.08 \\

& 55.44 $\pm$ 1.00&89.43 $\pm$ 0.25& 97.69 $\pm$ 0.06\\
     \midrule
 \multirow{3}{*}{LSUN-Crop}  &\cellcolor[gray]{0.8} 43.80 $\pm$ 1.29 &\cellcolor[gray]{0.8} 93.04 $\pm$ 0.22 &\cellcolor[gray]{0.8}98.56 $\pm$ 0.05   \\
 & \cellcolor[gray]{0.9}  87.32 $\pm$ 0.36 & \cellcolor[gray]{0.9} 83.09 $\pm$ 0.20 & \cellcolor[gray]{0.9} 96.40 $\pm$ 0.05 \\
 &74.89  $\pm$ 0.73  &84.98 $\pm$ 0.18& 96.46 $\pm$ 0.08\\
     \midrule
 \multirow{3}{*}{LSUN-Resize  } &\cellcolor[gray]{0.8} 71.86 $\pm$ 0.69 &\cellcolor[gray]{0.8} 81.86 $\pm$ 0.27&\cellcolor[gray]{0.8} 95.60 $\pm$ 0.09 \\
 & \cellcolor[gray]{0.9}   81.81 $\pm$ 0.71  &  \cellcolor[gray]{0.9 } 72.96 $\pm$  0.59 & \cellcolor[gray]{0.9} 91.64 $\pm$ 0.23\\
 &64.35 $\pm$ 0.62 & 82.64 $\pm$ 0.36 & 95.27 $\pm$ 0.14\\
 \midrule
 
  \multirow{3}{*}{\textbf{Mean} }  &\cellcolor[gray]{0.8} 73.11  &\cellcolor[gray]{0.8} 82.75 &\cellcolor[gray]{0.8}  \textbf{95.90} \\
  & \cellcolor[gray]{0.9} 76.62& \cellcolor[gray]{0.9} 77.37& \cellcolor[gray]{0.9} 93.69
\\
  & \textbf{67.99}  &  \textbf{83.17}& 95.70 \\
  
 \bottomrule
\end{tabular}}
\label{tab:cifar_100_ood_energy}
\end{table}

\section{The Effect of gradient update of Euclidean parameters on the hyperbolic embedding}
\label{sec: effect_of_update}

We derive the effect of the a single gradient update of the Euclidean parameters on the hyperbolic embedding. For the Euclidean sub-network $E: \mathbb{R}^m \rightarrow \mathbb{R}^n$. Consider the first-order Taylor-expansion of the Euclidean network with a single gradient update,
\begin{equation}
\begin{split}
 E(\mathbf{w}^E_{t+1}) & = E(\mathbf{w}_t^E + \eta \frac{\partial \ell}{\partial \mathbf{w}^E})  \\
 & \approx   E(\mathbf{w}^E_{t}) +   \eta (\frac{\partial E(\mathbf{w}_t^E)}{\partial \mathbf{w}_t^E})^T \frac{\partial \ell}{\partial \mathbf{w}^E} \\
 \end{split}
\end{equation}

Meanwhile, the exponentional map of the Poincar\'e ball is,

\begin{equation}
    \textnormal{Exp}_\mathbf{0}^c(\mathbf{v}) = \tanh (\sqrt{c} \lVert  \mathbf{v} \rVert) \frac{\mathbf{v
    }}{\sqrt{c} \lVert  \mathbf{v} \rVert}
\end{equation}

The gradient of the exponential map can be computed as,
\begin{equation}
\scriptsize
\begin{split}
    \nabla \textnormal{Exp}_\mathbf{0}^c(\mathbf{v}) & = \frac{\mathbf{v}}{\sqrt{c} \lVert \mathbf{v} \rVert} \nabla \tanh({\sqrt{c}\lVert \mathbf{v} \rVert}) + \tanh({\sqrt{c}\lVert \mathbf{v} \rVert}) \nabla \frac{\mathbf{v}}{\sqrt{c} \lVert  \mathbf{v} \rVert} \\
    & = 1-\tanh^2({\sqrt{c}\lVert \mathbf{v} \rVert}) + \tanh({\sqrt{c}\lVert \mathbf{v} \rVert})\frac{1}{\sqrt{c}}  \frac{2}{\lVert \mathbf{v} \rVert}
\end{split}
\end{equation}

Let $\mathbf{x}_{t+1}^H$ be the projected point in hyperbolic space, i.e, 
\begin{equation}
    \mathbf{x}_{t+1}^H = \textnormal{Exp}_\mathbf{0}^c(E(\mathbf{w}^E_{t+1}))
\end{equation}

Again we can apply the first-order Taylor-expansion on the exponential map,

\begin{equation}
\begin{split}
     \mathbf{x}_{t+1}^H & = \textnormal{Exp}_\mathbf{0}^c(E(\mathbf{w}^E_{t+1})) \\
            & \approx \textnormal{Exp}_\mathbf{0}^c(E(\mathbf{w}^E_{t}) + \eta (\frac{\partial E(\mathbf{w}_t^E)}{\partial \mathbf{w}_t^E})^T \frac{\partial \ell}{\partial \mathbf{w}^E}) \\
\end{split}
\end{equation}

Denote $\eta (\frac{\partial E(\mathbf{w}_t^E)}{\partial \mathbf{w}_t^E})^T \frac{\partial \ell}{\partial \mathbf{w}^E}$ by $J_{\mathbf{w}^E_t}$, we have

\begin{equation}
\begin{split}
     \mathbf{x}_{t+1}^H & = \textnormal{Exp}_\mathbf{0}^c(E(\mathbf{w}^E_{t+1})) \\
            & \approx \textnormal{Exp}_\mathbf{0}^c(E(\mathbf{w}^E_{t}) + J_{\mathbf{w}^E_t}) \\
            & \approx \textnormal{Exp}_\mathbf{0}^c(E(\mathbf{w}^E_{t})) + (\frac{\partial \textnormal{Exp}_\mathbf{0}^c(E(\mathbf{w}^E_{t}))}{\partial E(\mathbf{w}^E_{t})})^T J_{\mathbf{w}^E_t} \\
            & = \mathbf{x}_{t}^H + (\frac{\partial \textnormal{Exp}_\mathbf{0}^c(E(\mathbf{w}^E_{t}))}{\partial E(\mathbf{w}^E_{t})})^T J_{\mathbf{w}^E_t}
\end{split}
\end{equation}

Denote $(\frac{\partial \textnormal{Exp}_\mathbf{0}^c(E(\mathbf{w}^E_{t}))}{\partial E(\mathbf{w}^E_{t})})^T\eta (\frac{\partial E(\mathbf{w}_t^E)}{\partial \mathbf{w}_t^E})^T$ by $C(E(\mathbf{w}^E_{t}))$,

\begin{equation}
\begin{split}
     \mathbf{x}_{t+1}^H & = \mathbf{x}_{t}^H + C(E(\mathbf{w}^E_{t}))^T\frac{\partial \ell}{\partial \mathbf{w}^E}  \\
\end{split}
\end{equation}

\section{Datasets}
\label{sec:datasets}
The statistics of the datasets are shown in Table \ref{table: datasets_stat}.
\begin{table}[!ht]
\tiny
\renewcommand{\arraystretch}{0.6}
\setlength{\tabcolsep}{1pt}
\center
\small
\begin{tabular}{ l c c c c} 
\toprule
  & \textbf{MNIST} & \textbf{CIFAR10}  &  \textbf{CIFAR100}  & \textbf{ImageNet}  \\
\toprule
 \# of Training Examples &  60,000 &  50,000  & 50,000  &  1,281,167 \\
  \# of Test Examples &  10,000 &   10,000 &  10,000 &  50,000 \\
\hline 
\end{tabular}
\vspace{5pt}
\caption{The statistics of the datasets.}
\label{table: datasets_stat}
\end{table}

\section{The effect of hyperparameter r}
\label{sec:ablation_on_r}
We conduct ablation studies to show the effect of the hyperparameter $r$ which is the maximum norm of the Euclidean embedding. In Figure \ref{fig:ablation_on_r} we show the change of test accuracy as we vary the hyperparameter $r$ on MNIST, CIFAR10 and CIFAR100. We repeat the experiments for each choice of $r$ five times and report both average accuracy and standard deviation. On the one hand, it can be observed that a larger $r$ leads to a drop in test accuracy. As we point out, this is caused by the vanishing gradient problem in training hyperbolic neural networks. On the other hand, a small $r$ can also lead to a drop in test accuracy especially on more complex tasks such as CIFAR10 and CIFAR100. The plausible reason is that a small $r$ reduces the capacity of the embedding space which is detrimental for learning discriminative features. 

To conclude, there is a sweet spot in terms of choosing $r$ which is neither too large (causing vanishing gradient problem) nor too small (not enough capacity). The performance of clipped HNNs is also robust to the choice of the hyperparameter $r$ if it is around the sweet spot.

\begin{figure*}[!t]
    \centering
    \includegraphics[width=0.9\textwidth]{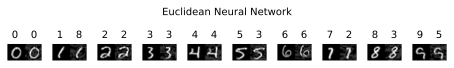}
    \includegraphics[width=0.9\textwidth]{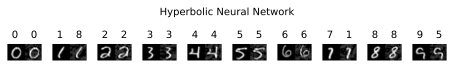}
    \caption{Clipped HNNs show more adversarial robustness compared with ENNs. We show the clean image and the corresponding adversarial image and the predictions of the network of 10 randomly sampled images. In several cases, clipped HNNs make correct predictions on the adversarial images while Euclidean neural networks make wrong predictions. }
    \label{figs: clean_img_adv_img}
\end{figure*}

\section{More results on adversarial robustness}
\label{sec:more_on_adv}
Although we observe that with adversarial training, hyperbolic neural networks achieve similar robust accuracy to Euclidean neural networks, in a further study, we consider training models using a small $\epsilon$ but attacking with a larger $\epsilon$ with FGSM on MNIST. In Table \ref{table: train_small_attack_large} we show the results of training the networks using $\epsilon = 0.05$ and attacking with $\epsilon $ = 0.1, 0.2 and 0.3. We can observe that for attacking with larger $\epsilon$ such as 0.2 and 0.3, clipped HNNs show more robustness to ENNs. Moreover, clipped HNNs greatly improve the robustness of vanilla HNNs. The possible explanation is that the proposed feature clipping reduces the adversarial noises in the forward pass and also improve the performance of vanilla HNNs. One of the future directions is to systematically understand and analyze the reason behind the robustness of clipped HNNs. In Figure \ref{figs: clean_img_adv_img}, we show the clean and adversarial images generated by FGSM with clipped HNNs and ENNs respectively. The predictions of the networks are shown above the image. It can be observed that clipped HNNs show more adversarial robustness compared with ENNs.

\begin{table}[!ht]
\center
\scriptsize
\scalebox{0.9}{
\begin{tabular}{ c|c|c|c} 
 \toprule
 \diagbox{Network}{$\epsilon$} &   0.1  &  0.2  &  0.3 \\
\toprule
ENNs & \cellcolor[gray]{0.9} 94.51 $\pm$ 0.32 \% & \cellcolor[gray]{0.9} 67.85 $\pm$ 2.12 \%  &  \cellcolor[gray]{0.9} 42.18 $\pm$ 1.32 \% \\
\midrule
Vanilla HNNs  & 81.08 $\pm$ 4.03 \% & 46.57 $\pm$ 2.09 \%  & 17.21 $\pm$ 3.27\%   \\
\midrule
Clipped HNNs &   93.34 $\pm$ 0.16 \% &  74.97 $\pm$ 1.02 \% &   46.27 $\pm$ 1.88 \% \\
\bottomrule
\end{tabular}
}
\caption{Adversarial training with FGSM ($\epsilon$ = 0.05) on MNIST. For attacking with larger $\epsilon$ such as 0.1, 0.2 and 0.3, clipped HNNs greatly improve the robustness of vanilla HNNs and show more robustness to ENNs when the attacking with large perturbations ($\epsilon$  = 0.2 and $\epsilon$ = 0.3).}
\label{table: train_small_attack_large}
\end{table}

\section{More results on out-of-distribution detection (OOD)}
\label{sec:more_on_ood}

\subsection{Results with energy score}
In Table \ref{tab:cifar_10_ood_energy} and \ref{tab:cifar_100_ood_energy} we show the results of using energy score \cite{liu2020energy} on CIFAR10 and CIFAR100 for out-of-distribution detection. We can observe that on CIFAR10, clipped HNNs achieve comparable performance in terms of FPR95 and perform much better in terms other AUROC and AUPR compared with ENNs. On CIFAR100, clipped HNNs achieve comparable performance in terms of AUPT and perform much better in terms other FPR95 and AUROC compared with ENNs. The results are consistent with the case of using softmax score.

\subsection{Clipped HNNs greatly improve the OOD detection capability of vanilla HNNs.}
In Table \ref{tab:cifar_10_ood_softmax_s} - Table \ref{tab:cifar_100_ood_energy}, we also show the results of using vanilla HNNs for out-of-distribution detection. Across all the datasets and scores, we can see that clipped HNNs greatly improve the OOD detection capability of vanilla HNNs. This shows that vanilla HNNs have poor OOD detection ability which can greatly limit their practical applicability.

\begin{figure*}[!ht]
   \centering
\begin{tabular}{cc}
\includegraphics[width=0.4\textwidth]{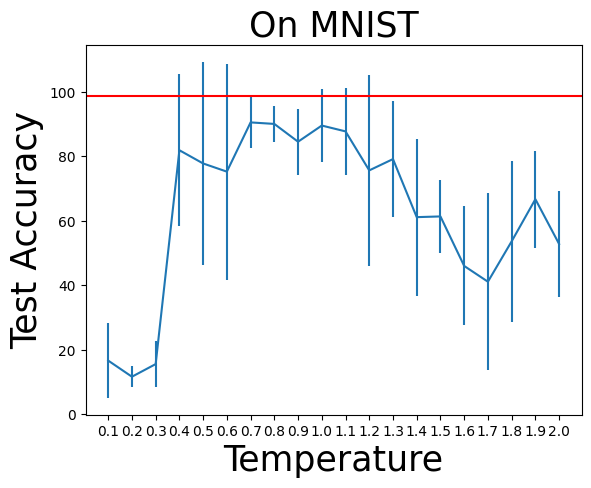}& 
\includegraphics[width=0.4\textwidth]{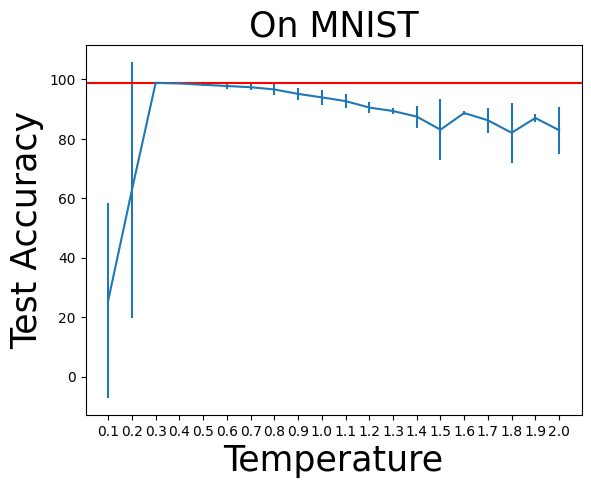}
\end{tabular}
    \caption{We show the change of the test accuracy as we vary the temperature parameter $T$. The red horizontal line is the result of the hyperbolic neural networks with the proposed feature clipping. Softmax with temperature scaling with a carefully tuned temperature can approach the performance of the proposed feature clipping. However, it is sensitive to the feature dimension and the temperature parameter.
    \textbf{Left}: the embedding dimension is 2. \textbf{Right}: the embedding dimension is 64. }
     \label{fig:ablation_on_t}
\end{figure*}

\section{Softmax with temperature scaling}
\label{sec:temperature}
We consider softmax with temperature scaling as an alternative for addressing the vanishing gradient problem in training hyperbolic neural networks. Softmax with temperature scaling introduces an additional temperature parameter $T$ to adjust the logits before applying the softmax function. Softmax with temperature scaling can be formulated as,

\begin{center}
\begin{equation}
\scriptsize
    \textnormal{Softmax}(\bm{Z} / T)_i = \frac{e^{Z_i / T}}{ \sum_{j=1}^K e^{Z_j / T}}  \quad \textnormal{for} \quad i = 1, ..., K, \quad \bm{Z} = (Z_1, ..., Z_K)
    \label{eq:softmax}
\end{equation}
\end{center}

In hyperbolic neural networks, $\bm{Z}$ is the output of the hyperbolic fully-connected layer and $K$ is the number of classes. If the additional temperature parameter $T$ is smaller than 1, the magnitude (in the Euclidean sense) of the hyperbolic embedding will be scaled up which prevents it from approaching the boundary of the ball.

In Figure \ref{fig:ablation_on_t}, we show the performance of training hyperbolic neural networks with temperature scaling compared with the proposed feature clipping. We consider feature dimensions of 2 and 64 respectively. Different temperature parameters are considered and the experiments are repeated for 10 times with different random seeds. We show both the average accuracy and the standard deviation. We can observe that softmax with temperature scaling and a carefully tuned temperature parameter can approach the performance of the proposed feature clipping when the feature dimension is 2. However, the feature dimension is 64, softmax with temperature scaling severely underperforms the proposed feature clipping. The results again confirm the effectiveness of the proposed approach.

\begin{figure*}[!ht]
   \centering
\begin{tabular}{cc}
\includegraphics[width=0.4\textwidth]{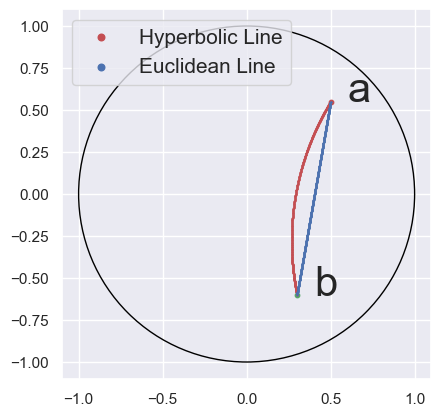}& 
\includegraphics[width=0.4\textwidth]{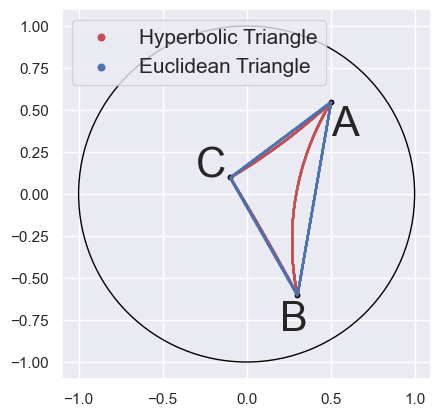}
\end{tabular}
    \caption{A magnitude-clipped hyperbolic space is still hyperbolic and behaves drastically different from Euclidean space. \textbf{Left:} the comparison of hyperbolic line segment with Euclidean line segment.  \textbf{Right:} the comparison of hyperbolic triangle with Euclidean triangle. }
     \label{fig:still_hyperbolic}
\end{figure*}

\section{A magnitude-clipped hyperbolic space is still hyperbolic }
\label{sec:still_hyperbolic}
 The metric in the hyperbolic space with the clipping strategy is still drastically different from that in the Euclidean space, even with magnitude clipping. For the first example, consider two points: $a$ = [0.5, 0.55], $b$ = [0.3, -0.6], the magnitude of both points is smaller than 0.76. The hyperbolic distance between the two points is 3.1822 while the Euclidean distance is 1.1673. This is a two-dimensional example, with a larger embedding dimension, the difference will be much more significant. In Figure \ref{fig:still_hyperbolic}: left, we compare the hyperbolic line segment  with Euclidean line segment given the point $a$ and the point $b$.

A magnitude-clipped hyperbolic space is still hyperbolic, as the hyperbolic geometry still holds: unlike Euclidean triangles, where the angles always add up to $\pi$ radians (180$^{\circ}$, a straight angle), in hyperbolic geometry the sum of the angles of a hyperbolic triangle is always strictly less than $\pi$ radians (180$^{\circ}$, a straight angle). The difference is referred to as the defect. For a second example, consider three points: A = [0.5, 0.55], B = [0.3, -0.6], C = [-0.1, 0.1]. Their magnitude are all smaller than 0.76. For the triangle ABC, the defect is 58.21$^{\circ}$ in hyperbolic space and 0$^{\circ}$ in Euclidean space. This again shows that the clipped hyperbolic space still well maintains the hyperbolic property. In Figure \ref{fig:still_hyperbolic}: right, we compare the hyperbolic triangle with Euclidean triangle given the point A, the point B and the point C.

We apply the proposed clipping strategy to learn word embedding as in \cite{nickel2017poincar}. We perform the reconstruction task on the transitive closure of the WordNet noun hierarchy. We compare the embedding quality of the Euclidean space, the hyperbolic space, clipped hyperbolic space using mean average precision (mAP). The embedding dimension is 10. The results are summarized in Table \ref{tab:word_embeddings}.

We have two conclusions here. First, learning word embeddings with hyperbolic space provides better results than learning in Euclidean space. Second, using hyperbolic space with clipping is slightly better than using hyperbolic space without clipping. 

\begin{table}[!ht]
    \centering
    \begin{tabular}{|c|c|}
    \toprule
        \textbf{Method}  &  \textbf{mAP} \\
       \midrule
       Euclidean space  &  	0.059 \\
       \midrule
       Clipped hyperbolic space & 	0.860\\
       \midrule
       Hyperbolic space  & 	0.851\\
       \bottomrule
    \end{tabular}
    \caption{Learning with word embeddings with clipped hyperbolic space outperforms both with Euclidean space and vanilla hyperbolic space.  }
    \label{tab:word_embeddings}
\end{table}

\section{Additional regularization to minimize the norm of the Euclidean embedding during training}
\label{sec:with_regularization}
The results of using the regularization term are shown in Table \ref{tab:regularization}.

\begin{table}[!ht]
    \centering
    \begin{tabular}{|c|c|c|}
    \toprule
        \textbf{Method} &   \textbf{On CIFR10} &  \textbf{On CIFAR100} \\
    \toprule
       vanilla HNN  &  	88.82 & 72.26 \\
       \midrule
       w/ regularization  & 	92.71 & 73.34 \\
\midrule
       w/ clipping  & 	94.76 & 75.88\\
       \bottomrule
    \end{tabular}
    \caption{The proposed feature clipping outperforms vanilla HNNs and HNNs with regularization. }
    \label{tab:regularization}
\end{table}
We can see that the clipping strategy outperforms the regularization approach. The reason is that with regularization, the loss function consists of two terms: one is the cross-entropy loss and the other is the regularization loss. It is difficult to balance the two terms . During training, if the cross-entropy loss becomes small, the optimization focuses on minimizing the embedding norm, however a small embedding norm is also detrimental to the performance.

\section{More discussions on Lorentz model  }
\label{sec:Lorentz}
Lorentz model is another commonly used model for hyperbolic space. The Lorentz model of $n$-dimensional hyperbolic space is defined as,

\begin{equation}
    \mathcal{H}^n = \{\mathbf{x} \in \mathbb{R}^{n+1}: \langle \mathbf{x}, \mathbf{x} \rangle_{\mathcal{L}} = -1 , x_0  > 0 \}
\end{equation}
where $\langle \mathbf{x}, \mathbf{x} \rangle_{\mathcal{L}}$ is the \emph{Lorentzian scalar product} which is defined as,

\begin{equation}
    \langle \mathbf{x}, \mathbf{y} \rangle_{\mathcal{L}} = -x_0y_0 + \sum_{i=1}^n x_iy_i
\end{equation}
The distance function of Lorentz model is given as,
\begin{equation}
    d_{\mathcal{H}}(\mathbf{x}, \mathbf{y}) = \arccosh(- \langle \mathbf{x}, \mathbf{y} \rangle_{\mathcal{L}}) 
\end{equation}

Lorentz model is used recently to overcome the numerical issues of the distance function in Poincar\'e ball model for learning word embeddings \cite{nickel2018learning}. However, it is most effective only in low dimensions \cite{liu2019hyperbolic}. For image datasets of ImageNet-scale, hyperbolic neural networks with high-dimensional embeddings are necessary for enough model capacity. Moreover, current hyperbolic neural network layers are only designed for Poinar\'e ball model. To extend hyperbolic neural networks layers for Lorentz model can be an interesting future work.

{\small
\bibliographystyle{ieee_fullname}
\bibliography{egbib}

\begin{thebibliography}{10}\itemsep=-1pt

\bibitem{alanis2016efficient}
Gregorio Alanis-Lobato, Pablo Mier, and Miguel~A Andrade-Navarro.
\newblock Efficient embedding of complex networks to hyperbolic space via their
  laplacian.
\newblock {\em Scientific reports}, 6(1):1--10, 2016.

\bibitem{anderson2006hyperbolic}
James~W Anderson.
\newblock {\em Hyperbolic geometry}.
\newblock Springer Science \& Business Media, 2006.

\bibitem{bonnabel2013stochastic}
Silvere Bonnabel.
\newblock Stochastic gradient descent on riemannian manifolds.
\newblock {\em IEEE Transactions on Automatic Control}, 58(9):2217--2229, 2013.

\bibitem{carmo1992riemannian}
Manfredo Perdigao~do Carmo.
\newblock {\em Riemannian geometry}.
\newblock Birkh{\"a}user, 1992.

\bibitem{cho2019large}
Hyunghoon Cho, Benjamin DeMeo, Jian Peng, and Bonnie Berger.
\newblock Large-margin classification in hyperbolic space.
\newblock In {\em The 22nd International Conference on Artificial Intelligence
  and Statistics}, pages 1832--1840. PMLR, 2019.

\bibitem{cimpoi2014describing}
Mircea Cimpoi, Subhransu Maji, Iasonas Kokkinos, Sammy Mohamed, and Andrea
  Vedaldi.
\newblock Describing textures in the wild.
\newblock In {\em Proceedings of the IEEE Conference on Computer Vision and
  Pattern Recognition}, pages 3606--3613, 2014.

\bibitem{deng2009imagenet}
Jia Deng, Wei Dong, Richard Socher, Li-Jia Li, Kai Li, and Li Fei-Fei.
\newblock Imagenet: A large-scale hierarchical image database.
\newblock In {\em 2009 IEEE conference on computer vision and pattern
  recognition}, pages 248--255. Ieee, 2009.

\bibitem{ganea2018hyperbolic}
Octavian-Eugen Ganea, Gary B{\'e}cigneul, and Thomas Hofmann.
\newblock Hyperbolic neural networks.
\newblock {\em arXiv preprint arXiv:1805.09112}, 2018.

\bibitem{goodfellow2014explaining}
Ian~J Goodfellow, Jonathon Shlens, and Christian Szegedy.
\newblock Explaining and harnessing adversarial examples.
\newblock {\em arXiv preprint arXiv:1412.6572}, 2014.

\bibitem{gulcehre2018hyperbolic}
Caglar Gulcehre, Misha Denil, Mateusz Malinowski, Ali Razavi, Razvan Pascanu,
  Karl~Moritz Hermann, Peter Battaglia, Victor Bapst, David Raposo, Adam
  Santoro, et~al.
\newblock Hyperbolic attention networks.
\newblock {\em arXiv preprint arXiv:1805.09786}, 2018.

\bibitem{gupte2011finding}
Mangesh Gupte, Pravin Shankar, Jing Li, Shanmugauelayut Muthukrishnan, and
  Liviu Iftode.
\newblock Finding hierarchy in directed online social networks.
\newblock In {\em Proceedings of the 20th international conference on World
  wide web}, pages 557--566, 2011.

\bibitem{hanin2018neural}
Boris Hanin.
\newblock Which neural net architectures give rise to exploding and vanishing
  gradients?
\newblock {\em arXiv preprint arXiv:1801.03744}, 2018.

\bibitem{hazan2015beyond}
Elad Hazan, Kfir~Y Levy, and Shai Shalev-Shwartz.
\newblock Beyond convexity: Stochastic quasi-convex optimization.
\newblock {\em arXiv preprint arXiv:1507.02030}, 2015.

\bibitem{he2016deep}
Kaiming He, Xiangyu Zhang, Shaoqing Ren, and Jian Sun.
\newblock Deep residual learning for image recognition.
\newblock In {\em Proceedings of the IEEE conference on computer vision and
  pattern recognition}, pages 770--778, 2016.

\bibitem{hochreiter1998vanishing}
Sepp Hochreiter.
\newblock The vanishing gradient problem during learning recurrent neural nets
  and problem solutions.
\newblock {\em International Journal of Uncertainty, Fuzziness and
  Knowledge-Based Systems}, 6(02):107--116, 1998.

\bibitem{hochreiter1997long}
Sepp Hochreiter and J{\"u}rgen Schmidhuber.
\newblock Long short-term memory.
\newblock {\em Neural computation}, 9(8):1735--1780, 1997.

\bibitem{hsu2020learning}
Joy Hsu, Jeffrey Gu, Gong-Her Wu, Wah Chiu, and Serena Yeung.
\newblock Learning hyperbolic representations for unsupervised 3d segmentation.
\newblock {\em arXiv preprint arXiv:2012.01644}, 2020.

\bibitem{khrulkov2020hyperbolic}
Valentin Khrulkov, Leyla Mirvakhabova, Evgeniya Ustinova, Ivan Oseledets, and
  Victor Lempitsky.
\newblock Hyperbolic image embeddings.
\newblock In {\em Proceedings of the IEEE/CVF Conference on Computer Vision and
  Pattern Recognition}, pages 6418--6428, 2020.

\bibitem{kingma2013auto}
Diederik~P Kingma and Max Welling.
\newblock Auto-encoding variational bayes.
\newblock {\em arXiv preprint arXiv:1312.6114}, 2013.

\bibitem{klimovskaia2020poincare}
Anna Klimovskaia, David Lopez-Paz, L{\'e}on Bottou, and Maximilian Nickel.
\newblock Poincar{\'e} maps for analyzing complex hierarchies in single-cell
  data.
\newblock {\em Nature communications}, 11(1):1--9, 2020.

\bibitem{krizhevsky2009learning}
Alex Krizhevsky, Geoffrey Hinton, et~al.
\newblock Learning multiple layers of features from tiny images.
\newblock 2009.

\bibitem{lecun1998mnist}
Yann LeCun.
\newblock The mnist database of handwritten digits.
\newblock {\em http://yann. lecun. com/exdb/mnist/}, 1998.

\bibitem{lecun1998gradient}
Yann LeCun, L{\'e}on Bottou, Yoshua Bengio, and Patrick Haffner.
\newblock Gradient-based learning applied to document recognition.
\newblock {\em Proceedings of the IEEE}, 86(11):2278--2324, 1998.

\bibitem{lee2018introduction}
John~M Lee.
\newblock {\em Introduction to Riemannian manifolds}.
\newblock Springer, 2018.

\bibitem{liu2019hyperbolic}
Qi Liu, Maximilian Nickel, and Douwe Kiela.
\newblock Hyperbolic graph neural networks.
\newblock {\em arXiv preprint arXiv:1910.12892}, 2019.

\bibitem{liu2020energy}
Weitang Liu, Xiaoyun Wang, John~D Owens, and Yixuan Li.
\newblock Energy-based out-of-distribution detection.
\newblock {\em arXiv preprint arXiv:2010.03759}, 2020.

\bibitem{loshchilov2016sgdr}
Ilya Loshchilov and Frank Hutter.
\newblock Sgdr: Stochastic gradient descent with warm restarts.
\newblock {\em arXiv preprint arXiv:1608.03983}, 2016.

\bibitem{madry2017towards}
Aleksander Madry, Aleksandar Makelov, Ludwig Schmidt, Dimitris Tsipras, and
  Adrian Vladu.
\newblock Towards deep learning models resistant to adversarial attacks.
\newblock {\em arXiv preprint arXiv:1706.06083}, 2017.

\bibitem{mathieu2019continuous}
Emile Mathieu, Charline~Le Lan, Chris~J Maddison, Ryota Tomioka, and Yee~Whye
  Teh.
\newblock Continuous hierarchical representations with poincar$\backslash$'e
  variational auto-encoders.
\newblock {\em arXiv preprint arXiv:1901.06033}, 2019.

\bibitem{miller1995wordnet}
George~A Miller.
\newblock Wordnet: a lexical database for english.
\newblock {\em Communications of the ACM}, 38(11):39--41, 1995.

\bibitem{mishkin2015all}
Dmytro Mishkin and Jiri Matas.
\newblock All you need is a good init.
\newblock {\em arXiv preprint arXiv:1511.06422}, 2015.

\bibitem{nagano2019wrapped}
Yoshihiro Nagano, Shoichiro Yamaguchi, Yasuhiro Fujita, and Masanori Koyama.
\newblock A wrapped normal distribution on hyperbolic space for gradient-based
  learning.
\newblock In {\em International Conference on Machine Learning}, pages
  4693--4702. PMLR, 2019.

\bibitem{netzer2011reading}
Yuval Netzer, Tao Wang, Adam Coates, Alessandro Bissacco, Bo Wu, and Andrew~Y
  Ng.
\newblock Reading digits in natural images with unsupervised feature learning.
\newblock 2011.

\bibitem{nickel2017poincar}
Maximilian Nickel and Douwe Kiela.
\newblock Poincar$\backslash$'e embeddings for learning hierarchical
  representations.
\newblock {\em arXiv preprint arXiv:1705.08039}, 2017.

\bibitem{nickel2018learning}
Maximillian Nickel and Douwe Kiela.
\newblock Learning continuous hierarchies in the lorentz model of hyperbolic
  geometry.
\newblock In {\em International Conference on Machine Learning}, pages
  3779--3788. PMLR, 2018.

\bibitem{pennington2018emergence}
Jeffrey Pennington, Samuel Schoenholz, and Surya Ganguli.
\newblock The emergence of spectral universality in deep networks.
\newblock In {\em International Conference on Artificial Intelligence and
  Statistics}, pages 1924--1932. PMLR, 2018.

\bibitem{pennington2017resurrecting}
Jeffrey Pennington, Samuel~S Schoenholz, and Surya Ganguli.
\newblock Resurrecting the sigmoid in deep learning through dynamical isometry:
  theory and practice.
\newblock {\em arXiv preprint arXiv:1711.04735}, 2017.

\bibitem{rumelhart1986learning}
David~E Rumelhart, Geoffrey~E Hinton, and Ronald~J Williams.
\newblock Learning representations by back-propagating errors.
\newblock {\em nature}, 323(6088):533--536, 1986.

\bibitem{russakovsky2015imagenet}
Olga Russakovsky, Jia Deng, Hao Su, Jonathan Krause, Sanjeev Satheesh, Sean Ma,
  Zhiheng Huang, Andrej Karpathy, Aditya Khosla, Michael Bernstein, et~al.
\newblock Imagenet large scale visual recognition challenge.
\newblock {\em International journal of computer vision}, 115(3):211--252,
  2015.

\bibitem{sala2018representation}
Frederic Sala, Chris De~Sa, Albert Gu, and Christopher R{\'e}.
\newblock Representation tradeoffs for hyperbolic embeddings.
\newblock In {\em International conference on machine learning}, pages
  4460--4469. PMLR, 2018.

\bibitem{sarkar2011low}
Rik Sarkar.
\newblock Low distortion delaunay embedding of trees in hyperbolic plane.
\newblock In {\em International Symposium on Graph Drawing}, pages 355--366.
  Springer, 2011.

\bibitem{shimizu2020hyperbolic}
Ryohei Shimizu, Yusuke Mukuta, and Tatsuya Harada.
\newblock Hyperbolic neural networks++.
\newblock {\em arXiv preprint arXiv:2006.08210}, 2020.

\bibitem{snell2017prototypical}
Jake Snell, Kevin Swersky, and Richard~S Zemel.
\newblock Prototypical networks for few-shot learning.
\newblock {\em arXiv preprint arXiv:1703.05175}, 2017.

\bibitem{ungar2001hyperbolic}
Abraham~A Ungar.
\newblock Hyperbolic trigonometry and its application in the poincar{\'e} ball
  model of hyperbolic geometry.
\newblock {\em Computers \& Mathematics with Applications}, 41(1-2):135--147,
  2001.

\bibitem{ungar2005analytic}
Abraham~A Ungar.
\newblock {\em Analytic hyperbolic geometry: Mathematical foundations and
  applications}.
\newblock World Scientific, 2005.

\bibitem{ungar2008gyrovector}
Abraham~Albert Ungar.
\newblock A gyrovector space approach to hyperbolic geometry.
\newblock {\em Synthesis Lectures on Mathematics and Statistics}, 1(1):1--194,
  2008.

\bibitem{weber2020robust}
Melanie Weber, Manzil Zaheer, Ankit~Singh Rawat, Aditya Menon, and Sanjiv
  Kumar.
\newblock Robust large-margin learning in hyperbolic space.
\newblock {\em arXiv preprint arXiv:2004.05465}, 2020.

\bibitem{WelinderEtal2010}
P. Welinder, S. Branson, T. Mita, C. Wah, F. Schroff, S. Belongie, and P.
  Perona.
\newblock {Caltech-UCSD Birds 200}.
\newblock Technical Report CNS-TR-2010-001, California Institute of Technology,
  2010.

\bibitem{xu2015empirical}
Bing Xu, Naiyan Wang, Tianqi Chen, and Mu Li.
\newblock Empirical evaluation of rectified activations in convolutional
  network.
\newblock {\em arXiv preprint arXiv:1505.00853}, 2015.

\bibitem{xu2015turkergaze}
Pingmei Xu, Krista~A Ehinger, Yinda Zhang, Adam Finkelstein, Sanjeev~R
  Kulkarni, and Jianxiong Xiao.
\newblock Turkergaze: Crowdsourcing saliency with webcam based eye tracking.
\newblock {\em arXiv preprint arXiv:1504.06755}, 2015.

\bibitem{yu2015lsun}
Fisher Yu, Ari Seff, Yinda Zhang, Shuran Song, Thomas Funkhouser, and Jianxiong
  Xiao.
\newblock Lsun: Construction of a large-scale image dataset using deep learning
  with humans in the loop.
\newblock {\em arXiv preprint arXiv:1506.03365}, 2015.

\bibitem{zagoruyko2016wide}
Sergey Zagoruyko and Nikos Komodakis.
\newblock Wide residual networks.
\newblock {\em arXiv preprint arXiv:1605.07146}, 2016.

\bibitem{zhang2019hyperbolic}
Yiding Zhang, Xiao Wang, Xunqiang Jiang, Chuan Shi, and Yanfang Ye.
\newblock Hyperbolic graph attention network.
\newblock {\em arXiv preprint arXiv:1912.03046}, 2019.

\bibitem{zhou2017places}
Bolei Zhou, Agata Lapedriza, Aditya Khosla, Aude Oliva, and Antonio Torralba.
\newblock Places: A 10 million image database for scene recognition.
\newblock {\em IEEE transactions on pattern analysis and machine intelligence},
  40(6):1452--1464, 2017.

\bibitem{zhou2020graph}
Jie Zhou, Ganqu Cui, Shengding Hu, Zhengyan Zhang, Cheng Yang, Zhiyuan Liu,
  Lifeng Wang, Changcheng Li, and Maosong Sun.
\newblock Graph neural networks: A review of methods and applications.
\newblock {\em AI Open}, 1:57--81, 2020.

\end{thebibliography}
}

\end{document}